\documentclass[cleveref]{alt2027} %

\usepackage{hyperref} 
\hypersetup{ hidelinks }%
\usepackage{amsfonts}   
\usepackage{nicefrac}  
\usepackage{enumitem}
\usepackage{stmaryrd}
\SetSymbolFont{stmry}{bold}{U}{stmry}{m}{n}
\usepackage{xspace}		%
\usepackage{booktabs} %
\usepackage{multirow}
\usepackage{mathtools}
\SetKwInput{KwInit}{Initialization}
\SetKwInput{KwDef}{Define}
\usepackage{thmtools,thm-restate}
\usepackage{bm}
\DeclareMathAlphabet{\mymathbb}{U}{BOONDOX-ds}{m}{n}
\usepackage{nicefrac}
\usepackage{etoolbox}
\usepackage{tikz}
\usepackage{mathrsfs}
\usepackage{etoolbox}
\AtBeginEnvironment{algorithm}{\setcounter{AlgoLine}{0}}

\usepackage{dutchcal}
\makeatletter
\providecommand*{\approxident}{%
  \mathrel{%
    \mathpalette\@approxident\sim
  }%
}   
\newcommand*{\@approxident}[2]{%
  \sbox0{$#1\vcenter{}$}%
  \sbox2{$\m@th#1\equiv$}%
  \dimen2=\dimexpr\ht2 - \ht0\relax
  \sbox4{$\m@th#1\sim$}%
  \dimen4=\dimexpr\ht4 - \ht0\relax
  \dimen0=\dimexpr
    -\ht4 - \dp4 %
    + \dimen2 %
  \relax
  \vcenter{\offinterlineskip
    \copy4 %
    \kern\dimen0 %
    \copy4 %
    \kern\dimen0 %
    \copy4 %
    \ifdim\dp4=\z@
      \kern\dimexpr -\ht0 + \dimen4\relax
    \fi
  }%
}      
\makeatother

\newtheorem{assumption}[theorem]{Assumption}

\crefname{condition}{condition}{conditions}

\usepackage{macros}

\renewcommand{\prtfun}{\parent}

\renewcommand{\prtat}[2]{\debug \prtfun(#1)}
\newcommand{\ancestor}{\debug {\mathfrak{an}}}
\renewcommand{\ancfun}{\ancestor}
\renewcommand{\anc}[2]{{ \ancfun_{#2}(#1)}}
\renewcommand{\avgnodeloss}{\debug {\expertloss}}
\renewcommand{\estnodeloss}{\debug {\hat{\expertloss}}}

\renewcommand{\cP}{{\debug{\Delta}}}
\renewcommand{\cF}{{\debug{\mathfrak{F}}}}
\renewcommand{\cnode}{\vertices}

\title[Contextual Non-Convex Bandit Optimization with Two-Point Feedback]{Contextual Non-Convex Bandit Optimization\\ with Two-Point Feedback}
\usepackage{times}

\altauthor{%
 \Name{Khaled Eldowa} \Email{khaled.eldowa@inria.fr}\\
 \addr Univ. Grenoble Alpes, 
 Inria, CNRS, Grenoble INP, LJK; Grenoble, France
 \AND
 \Name{Pierre Gaillard} \Email{pierre.gaillard@inria.fr}\\
 \addr Inria, D.I. École Normale Supérieure, CNRS, PSL Research University; Paris, France
}

\makeatletter
\IfFormatAtLeastTF{2024-06-01}{%
  \patchcmd{\@begintheorem}{\hskip\labelsep}
    {\MakeLinkTarget{\@currentcounter}\hskip\labelsep}{}{\PatchFailed}%
  \patchcmd{\@opargbegintheorem}{\hskip\labelsep}
    {\MakeLinkTarget{\@currentcounter}\hskip\labelsep}{}{\PatchFailed}%
}{}
\makeatother

\begin{document}
\maketitle

\begin{abstract}
We study a non-stochastic contextual bandit problem with Lipschitz losses. At every round, the learner observes the context then picks two actions, observes the difference between their losses, and incurs their mean. The aim is to minimize the regret against a generic non-parametric comparator class of policies mapping contexts to actions; for example, Lipschitz policies and H\"older-smooth policies. We show how relying on a hierarchical discretization technique allows us to improve upon the regret rates achievable under one-point feedback, and even recover the full-information rate in some cases; notably, for sufficiently smooth policy classes or low-dimensional instances. Additionally, for Lipschitz policies, we present an algorithm whose per-round computational complexity is only logarithmic in the time horizon.
\end{abstract}


\section{Introduction}

Many modern bandit problems are contextual and non-convex: at every round a decision maker observes side information (a context), must commit to an action in a possibly large and structured action set $\actions$, and only observes the losses of played actions. Examples include dynamic pricing, bidding and ad placement, adaptive control and hyperparameter tuning, where the quality of an action is only observable through a noisy or adversarial evaluation.
A natural benchmark in these settings is a rich, nonparametric class of policies $\experts$ mapping contexts to actions, for example all Lipschitz or H\"older-smooth functions. The price of considering such large nonparametric classes is a potentially slower sublinear regret rate, with the rate determined by the complexity of both the action space and the policy class, as quantified by their covering numbers and metric entropy, respectively.
The achievable regret rate depends strongly on the feedback model (see related work below). 
In this work, we are considering a setting, detailed below, that reveals %
more information about the loss function than standard bandit feedback. 

\paragraph{Setting.} Suppose we have an action set $\actions$, a context set $\contexts$, and a policy set $\experts$ consisting of functions mapping $\contexts$ to $\actions$.
Assume moreover that $\actions$ is equipped with a metric $\adist$ with respect to which it has a finite diameter.
Define the loss set $\losses$ as the set of $1$-Lipschitz functions mapping $\actions$ to $\R$.\footnote{For our purposes, one can also start with a generic loss set $\losses$ and take $\adist(a, b) \coloneqq \sup_{\loss \in \losses} |\loss(a) - \loss(b)|$ assuming the supremum is finite and non-zero for any pair of distinct actions.} 
We consider an oblivious environment that chooses at every round $t$ a context $x_t \in \contexts$ and a (possibly nonconvex) %
loss function $\loss_t \in \losses$.
After observing the context $x_t \in \contexts$, the learner selects two actions $A_{t,1}$ and $A_{t,2}$ in $\actions$, observes the difference between their losses $\loss_t(A_{t,1}) - \loss_t(A_{t,2})$, and suffers their average. The goal is to minimize the regret after $T$ rounds:
\begin{equation*}
\reg{T}{2} \coloneqq \E\left[\sum_{t=1}^T \frac{1}{2}\brb{ \loss_t(A_{t,1}) + \loss_t(A_{t,2}) }\right] - \inf_{\expert \in \experts}\ \sum_{t=1}^T \loss_t(\expert(x_t)) \,.
\end{equation*}
This model captures online decision problems in which a platform serves two users under the same market conditions and where both outcomes count towards performance.
In online advertising, for instance, the platform can show two users sharing the same context $x_t$ (query, time of day, etc.) two different ads (or bids)
$A_{t,1}$ and $A_{t,2}$.
The loss function $\loss_t$ encodes the latent market conditions (competing bids, demand, seasonality, etc.), which are common to both users
and possibly adversarial.
Similarly, a navigation app can recommend two alternative routes to two drivers with the same origin--destination pair and departure time, and observe their travel times (or their difference) under the same traffic conditions. 
In both examples, a reasonable performance measure is the average (or sum) of the losses of the two played actions.

\medskip
The central question of this work is to quantify the effect of two-point feedback in nonparametric nonconvex online learning: in particular when can the $\sqrt{T}$ rate of online convex optimization be recovered in this harder nonparametric bandit framework?

\paragraph{Contributions.} Our main contributions are as follows.
\begin{itemize}[leftmargin=0pt,
    labelsep=1em,
    labelwidth=.5em,
    itemindent=1.5em, topsep=5pt,parsep=0pt,itemsep=3pt]
\item \emph{A generic algorithm and regret bound (Section~\ref{sec:generic}).} We introduce an algorithm for contextual bandits with two-point feedback that operates over a generic tree of policy classes $(\expertree_\lvl)_{l=0}^{\nlvls}$, see \Cref{alg:2-point-feedback-experts}.
If the tree hierarchically approximates the policy class $\experts$ (in the sense of Assumption~\ref{assump:function-tree} below), then \Cref{thm:generic-thm} establishes the following regret bound:
\[
      \textstyle  \smash{ \reg{T}{2} \lesssim \Big( \sum_{\lvl} 2^{-2 \lvl / 3} \big(|\actions_{\lvl-1\to \lvl}| \cdot \ln |\expertree_\lvl| \big)^{1/3} \Big)^{3/2} \sqrt{T} + 2^{-\nlvls}T \,,}
\]
where $|\expertree_\lvl|$ is the number of policies at level $\lvl$ of the tree; and $|\actions_{\lvl-1\to \lvl}|  $ is the number of possible pairs of actions $(a_{\lvl - 1},a_{\lvl}) \in \actions^2$ such that $a_{\lvl -1}$ can be refined into $a_{\lvl}$ by a parent-child pair of functions in the policy tree between levels $\lvl-1$ and $\lvl$.
The generality of this result comes at a cost however: the bound takes an abstract form, and the corresponding algorithm is not computationally efficient in general.

\item \emph{Concrete rates (Section~\ref{sec:concrete_rate}).} For action spaces with covering dimension $\covdim$ and finite doubling dimension, and policy classes with metric-entropy exponent $\fundim$, we instantiate the construction of the policy tree and prove a regret of order
\begingroup
\setlength{\abovedisplayskip}{4pt} \setlength{\belowdisplayskip}{4pt}
\[
    \sqrt{T}\quad \text{when} \quad \covdim+\fundim < 2, \quad  \sqrt{T}(\ln T)^{3/2} \quad \text{when} \quad \covdim+\fundim = 2, \quad \text{and} \quad T^{1-1/(\covdim+\fundim)} \quad \text{otherwise}\,.
\] 
\endgroup
In particular, this is essentially the full-information rate when $\covdim+\fundim \le 2$, 
so two-point feedback closes the gap with full information in these regimes, and improves substantially over the best rates known under one-point feedback. Table~\ref{tab:rates} summarizes the known rates.
For instance, if $\actions \subseteq \R$, $\contexts \subseteq \R^\cdim$, and the policy set $\experts$ consists of $\holder$-Hölder smooth functions from $\contexts$ to $\actions$, we have $\fundim = {\contextdim}/{\holder}$ and $\adim=1$, which entails 
the full-information $\sqrt{T}$ rate as soon as the smoothness parameter $\holder$ is larger than the context dimension $\contextdim$.
Note that in the one-point bandit feedback framework, $\sqrt{T}$ is not achievable, even in the simplest setting when $\covdim = 1$ with no contexts (i.e., constant policies), where the minimax rate is $T^{2/3}$~\citep{kleinberg2008multi}. 

\item  \emph{A more efficient implementation for Lipschitz policies (Section~\ref{sec:efficient}).}  For the set of Lipschitz policies from $\contexts := [0,1]^\contextdim$ to $\actions := [0,1]^\covdim$, we provide a more efficient implementation (Algorithm~\ref{alg:efficient}) that is shown to be essentially equivalent to the generic Algorithm~\ref{alg:2-point-feedback-experts}. This is done by designing a policy tree with a separable structure, which allows merging known information on the past performance of different policies in a more compact format. 
In particular, for a fixed $\adim$, the overall space complexity of Algorithm~\ref{alg:efficient} is $\cO(T \ln T)$, while its per-round time complexity is only $\cO(\ln T)$. 
When $\contextdim = \covdim = 1$, our algorithm recovers the full-information $\sqrt{T}$ rate up to logarithmic factors but at a much lower computational cost compared to existing full-information procedures that are either non-constructive \citep{rakhlin2015onlinec} or require polynomial in $T$ per-round complexities to leverage the full-information feedback \citep{cesa2017algorithmic}.

\end{itemize}

\begin{table}[!t]
\small
\begin{center}
\begin{tabular}{llll} \toprule
& \bfseries Feedback model & \bfseries  Upper bound  & \bfseries Reference \\ \midrule
\multirow{3}{*}{\begin{tabular}[c]{@{}c@{}} Non-contextual \\ ($\fundim = 0)$ \end{tabular}} &      Bandit  & $\underline{T^{1-\frac{1}{\covdim+2}}}$ & \citep{kleinberg2008multi} \\[2pt]
   &   Two-point bandit  & $T^{1-\frac{1}{\covdim}} \lor \sqrt{T} $ & \citep{eldowa26a} \\[2pt] 
  &    Full information  & $\underline{\sqrt{T}} $ & \citep{rakhlin2015onlinec} \\[2pt]
  \hline & & \\[\dimexpr-\normalbaselineskip+2pt]
\multirow{4}{*}{\begin{tabular}[c]{@{}c@{}} Contextual \\ ($\fundim \geq 0)$ \end{tabular}}&      Bandit  & $T^{1-\frac{1}{\covdim+\fundim+2}}$ & \citep{slivkins2011contextual} \\[2pt]
   &   Two-point bandit  & $T^{1-\frac{1}{\covdim+\fundim}} \lor \sqrt{T} $ & {\bfseries (\Cref{thm:concrete})} \\[2pt] 
 &    One-sided full information ($\covdim = 1$) &  $T^{1-\frac{1}{\fundim+2/3}} \lor \sqrt{T}$ & \citep{cesa2017algorithmic} \\[2pt] 
  &    Full information  & $\underline{T^{1-\frac{1}{\fundim}} \lor \sqrt{T}} $ & \citep{rakhlin2015onlinec} \\ 
    \bottomrule
\end{tabular}
\end{center}
\caption{Regret bounds (up to logs) in nonparametric adversarial bandits with Lipschitz losses according to the level of feedback. 
The results of \cite{cesa2017algorithmic} and \cite{slivkins2011contextual} concern classes of Lipschitz policies.
Underlined rates are minimax optimal.
A matching lower bound for the contextual bandit setting was proved by \cite{slivkins2011contextual} when the composite losses $\loss_t(\func(\cdot))$ are assumed Lipschitz.
}
\label{tab:rates}
\end{table}

\paragraph{Relation to previous work.}
Our work builds on two lines of research: nonconvex nonparametric online learning, and multi-point feedback in bandit and zeroth-order optimization. 
Within the first line, we also discuss the common technique of building a hierarchical covering of the policy set.

\begin{itemize}[leftmargin=0pt,
    labelsep=1em,
    labelwidth=.5em,
    itemindent=1.5em, topsep=0pt,itemsep=5pt,parsep=0pt,topsep=4pt]
\item \emph{Nonconvex nonparametric online learning.} 
With \emph{full information}, 
i.e., when the entire loss function is revealed, 
most existing work considers convex losses. The case of generic Lipschitz loss is considered in \cite{hazan2007online,rakhlin2015onlinec,cesa2017algorithmic} and the rate is well understood.  For a policy class with metric entropy of order $\epsilon^{-p}$ (see \Cref{def:metric_entropy}), the minimax rate is $\sqrt{T}$ when $p \leq 2$ and $T^{(p-1)/p}$ when $p>2$ achieved by a nonconstructive procedure in \cite{rakhlin2015onlinec} and for Lipschitz policies with a polynomial algorithm in \cite{cesa2017algorithmic}.
The lower bound is due to \citet{hazan2007online}. With \emph{one-point bandit feedback}, where only the loss of the played action is revealed, the exponent degrades markedly; for Lipschitz losses and policies, the best known rates \citep{slivkins2011contextual,cesa2017algorithmic} are polynomially worse than their full-information counterparts.
In particular, when the covering number of the action set scales as $\epsilon^{-d}$ (see \Cref{def:covering}), the best known rate is $\tilde \cO(T^{\nicefrac{(d+p+1)}{(d+p+2)}})$. 

Recovering the full-information exponent under bandit feedback is impossible in general \citep{kleinberg2008multi,bubeck2011x}, and the gap between the two regimes is an obstacle to deploying nonparametric contextual methods in practice. In this work, we therefore consider the intermediate model of \emph{two-point bandit feedback} that has not been considered in nonparametric online learning. The intermediate feedback model of one-sided feedback was considered by \citet{cesa2017algorithmic} but it is a stronger feedback restricted to one-dimensional action sets and narrower applications (such as online auctions); moreover, they only considered Lipschitz classes of policies.

\item \emph{Hierarchical tree covering of the policy set.} A key technique in nonparametric online learning consists in building a tree that hierarchically approximates the policy set with increasing levels of precision. This was first used in online learning by \citet{cesa1999prediction} for linear loss functions, by \citet{rakhlin2014online,rakhlin2015onlineb} and \cite{rakhlin2015onlinec} to design nonconstructive procedures achieving the minimax rates in several nonparametric online frameworks, by \citet{gaillard2015chaining} to build a constructive minimax algorithm in online nonparametric regression, and later by  \citet{cesa2017algorithmic} for nonparametric bandits with one-sided feedback. 
An analogous technique was also used by \citet{eldowa26a} for (non-contextual) structured bandits with two-point feedback, where they construct a tree over the actions.

The key idea is that when the losses are Lipschitz, two nearby actions have a small loss difference, which, if observable, can allow building a loss estimator with small variance.
A tree structure where the nodes at successive levels cover the actions at decreasing scales provides a regular and simple representation of the metric (and hence the loss) structure, which can be exploited by adapting standard algorithmic frameworks.
The same applies when the decision space is one of policies instead of actions, as the distance between policies can be measured, for example, by the sup metric.
In this work, we leverage our access to loss differences and employ this hierarchical discretization idea to our particular setting of contextual bandit optimization with two-point feedback.
Importantly, we explore in the process the challenges (and achievable results) brought about when combining said technique with EXP4-style algorithms common in the contextual bandit space \citep{auer2002nonstochastic}.

Outside nonparametric problems, abstract bandit models with hierarchical action spaces are not uncommon in the literature. 
Most relevant here is the work of \cite{martin2022nested}, who relied on semi-bandit feedback to leverage the structure of a nested action set, and its expert-advice extension in \cite[Section 5.5]{zenati2023efficient}, where a finite and unstructured comparator class of experts was considered.

\item \emph{Two-point bandit feedback.} 
Multi-point bandit feedback has been considered in order to achieve improved performance compared to one-point bandit feedback. In (non-contextual) bandit convex optimization and zeroth-order optimization, a line of works \citep{agarwal2010optimal,duchi2015optimal,shamir2017optimal,lu2024adaptive} showed that two-point feedback, by reducing the variance of the gradient estimator, allows retrieving several benefits of the full-information setting, including a straightforward path to achieving the $O(\sqrt{T})$ minimax rate.
However, those results rely on convexity of the losses, on gradient estimators, 
and on having a parametric comparator.
None of this is available here: our losses are only Lipschitz (no convexity or differentiability assumed),
and the comparator set is a nonparametric class of functions. 
Similar improvements were shown in standard multi-armed bandits \citep{sani2014exploiting,thune2018adaptation,degenne2018}. However, in those works the extra observation is free, whereas here both played actions incur loss, which is more realistic in many applications.

Closest to this work is \cite{eldowa26a}, which considers a similar two-point feedback framework and shows that it allows leveraging the structure of the action set. In particular, they recover the $\sqrt{T}$ full-information rate in one or two-dimensional Lipschitz bandits. 
They however did not incorporate contexts and did not consider non-parametric policy classes. 
To our knowledge, the effect of two-point feedback in this regime has not been quantified.
\end{itemize}

\paragraph{Notation.}
For $v,w\in\R$, we use $v \lor w \coloneqq \max\{v,w\}$ and $v \land w \coloneqq \min\{v,w\}$.
For a positive integer $\cdim$, we use $[\cdim]$ to denote $\{1,\dots,\cdim\}$, $\ones_\cdim$ to denote the vector $(1,\ldots,1)^\top$ in $\R^\cdim$, and $\Delta_\cdim$ to denote the probability simplex in $\R^\cdim$.
We use $\indicator{\cdot}$ to denote the indicator function.
For a sequence of sets $(S_i)_{i\geq 1}$, we use $S_{1:n}$ to mean $\bigcup_{i=1}^n S_i$.

\section{A Generic Result}
\label{sec:generic}

In this section, we provide a generic algorithm that works on hierarchical trees that approximate the policy set. We first detail the tree approximation structure and the approximation properties that are required by our analysis and then present our algorithm together with its regret upper bound. 

\subsection{Hierarchical tree covering of the policy set}
\label{sec:tree}
Fix a positive integer $\nlvls$ and let $(\actions_\lvl)_{\lvl=0}^\nlvls$ be a sequence of finite subsets of actions in $\actions$ such that $\actions_0 \coloneqq \{a^*\}$ for some $a^* \in \actions$.
Assume we are given a sequence $(\functions_\lvl)_{\lvl=0}^\nlvls$ of finite sets of functions mapping $\contexts$ to $\actions$ such that $\functions_0 \coloneqq \{\func_0\}$ for some function $\func_0$, and $\image(\func) \subseteq \actions_\lvl$ for any $\func \in \functions_\lvl$.
Both of these sequences are allowed to contain overlapping sets.
We now use $(\functions_\lvl)_{\lvl=0}^\nlvls$ to build a tree of functions.
The set of nodes at level $\lvl$ is given by $\expertree_\lvl \coloneqq \{(\lvl,\func)\}_{\func \in \functions_\lvl}$; nevertheless, to declutter the notation, \emph{we refer to members of $\expertree_\lvl$ simply via the functions they represent}, and treat it as a set of functions in some statements.
In addition, we assume we are provided a parenthood function acting on all the nodes. 
In particular, the parent of $\func \in \expertree_\lvl$ is denoted by $\parent(\func) \in \expertree_{\lvl-1}$.
Building on this, an ancestor function can be defined over the nodes in an obvious manner:
$\anc{\func}{\lvl} \coloneqq \func$ for $\func \in \expertree_\lvl$ and $\anc{\func}{\lvl} \coloneqq \prtat{\anc{\func}{\lvl+1}}{\lvl}$ for $\func \in \expertree_{\lvl'}$ with $\lvl' > \lvl$.
We refer to the resulting tree of functions as $\cT(\expertree_{0:\nlvls})$.

We ask that $\cT(\expertree_{0:\nlvls})$ should possess two properties.
The first is that $\expertree_\nlvls$, the functions at the last level, approximate well the competitor space $\experts$, with a precision proportional to $2^{-\nlvls}$.
This would allow us to use $\expertree_\nlvls$ as a discrete surrogate for $\experts$.
The second property is that proximity in the function tree translates to similarity of behaviour. 
To make this concrete we define a notion of distance between functions: for $\func, \func'\colon \contexts \rightarrow \actions$, define 
\[    \textstyle
    \distfunc(\func,\func') \coloneqq \sup_{x \in \contexts} \adist(\func(x),\func'(x)) \,.\]
This is a metric over $\functionspace$, the set of all functions mapping $\contexts$ to $\actions$.
We can now state precisely the assumption that our algorithm will operate under.
\begin{assumption} \label{assump:function-tree}
    There exist two positive constants $\apprxcst$ and $\prtcst$ such that 
    \begin{enumerate}[nosep]
        \item $\expertree_\nlvls$ is a $\apprxcst 2^{-\nlvls}$-(external) cover of $\experts$ with respect to $\distfunc \,;$
        \item $\distfunc(\prtat{\func}{\lvl},\func) \leq \prtcst 2^{-\lvl}$ for 
        $\func \in \expertree_{\lvl+1} \,.$ 
    \end{enumerate}
\end{assumption}
To demonstrate the utility of the second property, observe that any $\func, \func' \in \expertree_\nlvls$ satisfying $\anc{\func}{\lvl} = \anc{\func'}{\lvl}$ also satisfy via the triangle inequality that
\begin{align*}
    \distfunc(\func,\func') &\leq \sum_{j = \lvl+1}^{\nlvls} \distfunc(\anc{\func}{j},\anc{\func}{j-1}) + \distfunc(\anc{\func'}{j},\anc{\func'}{j-1}) 
    \leq 4 \prtcst 2^{-\lvl } \,.
\end{align*}

The parenthood relation between the functions induces a relation between the actions which will emerge in the analyses to come.
Namely, we define the set of successors in $\actions_{\lvl+1}$ of an action $a \in \actions_\lvl$ as:
\begin{align} \label{def:succ-set}
    \suc_{\lvl+1}(a) \coloneqq \Big\{ b \in \actions_{\lvl+1} \mid \exists x \in \contexts ,\, \func \in \expertree_{\lvl+1}\:\:\mathrm{s.t.}\:\:
    a =  \parent(\func)(x) \:\: \text{and}  \:\: b = \func(x) \Big\} \,.
\end{align}
These sets might overlap for different actions in $\actions_\lvl$.
Notice now that the second property in \Cref{assump:function-tree} is equivalent to asking that for 
$a \in \actions_\lvl$ and $b \in \suc_{\lvl + 1}(a)$,
$
    \adist(a,b) \leq \prtcst 2^{-\lvl} \,.
$

\subsection{Algorithm and regret upper bound}
We are now ready to present \Cref{alg:2-point-feedback-experts}.
We use $\cP_{\expertree,\nlvls}$ to denote the space of probability distributions over $\expertree_\nlvls$;
also, $\pr(\cdot \mid \cF_{t-1})$ and $\E[\cdot \mid \cF_{t-1}]$ are denoted by $\pr_t(\cdot)$ and $\E_t[\cdot]$ respectively, where $\cF_{t}$ is the $\sigma$-algebra generated by the random events up to the end of round $t$. 
The algorithm is an instance of the follow-the-regularized-leader (FTRL) template executed on the leaves $\expertree_\nlvls$.
It builds upon Algorithm 1 of \cite{eldowa26a}, and in particular, extends its idea to the contextual (experts) setting via an EXP4-style estimator \citep{auer2002nonstochastic}.
Having observed $x_t$, the idea is to extend the cost function $\expertloss_t(\func) := \loss_t(\func(x_t))$ to the internal nodes of the tree, so that we can write $\expertloss_t(\func) = \sum_{\lvl \in [\nlvls]} \expertloss_{t,\lvl}(\anc{\func}{\lvl})$ up to an additive constant, see \Cref{lem:telescoping-nodeloss} in \Cref{app:generic-analysis}.
The value of $\expertloss_{t,\lvl}(\func')$ for some $\func' \in \expertree_\lvl$ is (up to a small constant shift) the expected value of the difference between the losses of two random actions:
the first is distributed as $\func(x_t)$ with $\func \sim p_t \mid \anc{\func}{\lvl}(x_t) = \func'(x_t)$, and the second as $\func(x_t)$ with $\func \sim p_t \mid \anc{\func}{\lvl-1}(x_t) = \parent(\func')(x_t)$.
Thanks to the second item in \Cref{assump:function-tree}, $\expertloss_{t,\lvl}(\func')$ is $\cO(2^{-\lvl})$.
Note that $\expertloss_{t,\lvl}(\func')$ provides information about all $\func \in \expertree_\nlvls$ with $\anc{\func}{\lvl}(x_t) = \func'(x_t)$ and $\anc{\func}{\lvl-1}(x_t) = \parent(\func')(x_t)$; that is, all the leaves $\func$ whose ancestors at levels $\lvl$ and $\lvl-1$ behave like, respectively, $\func'$ and $\parent(\func')$ on $x_t$.

With two-point feedback, we can compute an estimate of this value for one (randomly selected) %
node of the tree by sampling a pair of actions in the manner we just described.
The %
node selected in \Cref{alg:2-point-feedback-experts} is $\anc{\varfunc_t}{\lvl_t}$, and the two sampled actions are $A_{t,1} = \func_{t,1}(x_t)$ and $A_{t,2} = \func_{t,2}(x_t)$, whose distributions are consistent with our discussion above applied to the value of $\expertloss_{t,\lvl_t}(\anc{\varfunc_t}{\lvl_t})$.
This is then used to build an importance-weighted loss estimate $\hat{\expertloss}_t(\func)$ for all leaves $\func$, which is active only if $\anc{\func}{\lvl_t}$ and $\anc{\func}{\lvl_t-1}$ output the same actions as $\anc{\varfunc_{t}}{\lvl_t}$ and $\anc{\varfunc_{t}}{\lvl_t-1}$  at $x_t$.
Note that having $\varfunc_t \sim p_t$ guarantees that the expected loss of both $A_{t,1}$ and $A_{t,2}$ is the expected loss at round $t$ of a function sampled from $p_t$, see \Cref{lem:same-marginals-independent-from-level} in \Cref{app:generic-analysis}. 
Define for a distribution $p \in \cP_{\expertree,\nlvls}$ and $\func' \in \expertree_\lvl$ the quantity
$
    p[\func'] \coloneqq \sum_{\func \in \expertree_\nlvls} \indicator{ \func' = \anc{\func}{\lvl} } p(\func) 
$,
i.e., it is the total mass that $p$ places on the subtree rooted at $\func'$.
Using a nested negative entropy regularizer as in \cite{martin2022nested,eldowa26a}, we can achieve the following regret guarantee. (The analysis can be found in \Cref{app:generic-analysis}.)

\LinesNumbered
\begin{algorithm}[!t] 
 \KwIn{tree $\cT(\expertree_{0:\nlvls})$, regularizer $\psi$, constants $(b_\lvl)_{\lvl=1}^\nlvls$, distribution $\delta \in \Delta_{\nlvls}$}
 \KwInit{$p_1 = \argmin_{p \in \cP_{\expertree,\nlvls}} \psi(p)$}
 \For{$t \geq 1$}{

    sample $\varfunc_{t} \sim p_t$ and $\lvl_t \sim \delta$

    observe $x_t$\,;\, 
    set $\bee_{t,1} = \anc{\varfunc_{t}}{\lvl_t}(x_t)$ and $\bee_{t,2} = \anc{\varfunc_{t}}{\lvl_t-1}(x_t)$
    
    sample $\func_{t,1} \sim p_t \mid \anc{\func_{t,1}}{\lvl_t}(x_t) = \bee_{t,1}$ and $\func_{t,2} \sim p_t \mid \anc{\func_{t,2}}{\lvl_t-1}(x_t) = \bee_{t,2}$

    play $A_{t,1} = \func_{t,1}(x_t)$ and $A_{t,2} = \func_{t,2}(x_t)$\,;\, observe $\loss_t(A_{t,1}) - \loss_t(A_{t,2})$
    
    \vspace{0.1em}
    set 
    $\displaystyle
        \hat{\expertloss}_t(\func) = \frac{\indicator{  \bee_{t,1} = \anc{\func}{\lvl_t}(x_t) \,,\, \bee_{t,2}  = \anc{\func}{\lvl_t - 1}(x_t)  }}{\delta_{\lvl_t} \pr_t[ \bee_{t,1} = \anc{\func}{\lvl_t}(x_t) \,,\, \bee_{t,2}  = \anc{\func}{\lvl_t - 1}(x_t)    \mid \lvl_t]} \brb{\loss_t(A_{t,1}) - \loss_t(A_{t,2}) + b_{\lvl_t}}
    $
    
    \vspace{0.1em}
    set $\displaystyle p_{t+1} = \argmin_{p \in \cP_{\expertree,\nlvls}} \ban{\summ_{s \leq t} \hat{\expertloss}_s,p} + \psi(p)$
 }
 \caption{Policy-Tree FTRL with Two-Point Feedback} \label{alg:2-point-feedback-experts}
\end{algorithm}

\begin{theorem} \label{thm:generic-thm}
    Under \Cref{assump:function-tree},
    \Cref{alg:2-point-feedback-experts} run with $\psi(p) = \sum_{\lvl \in [\nlvls]} 1/\eta_{\lvl} \sum_{\func \in \expertree_\lvl} p[\func] \ln p[\func]$, $b_\lvl = \prtcst 2^{-\lvl+3}$, 
    $
        \eta_{\lvl} = \sqrt{ \frac{\delta_\lvl \ln |\expertree_\lvl|}{\prtcst^2 2^{-2\lvl+8} T |\actions_{\lvl-1\to\lvl}|} } \,,
        \:\text{ and }\:
        \delta_\lvl \propto 2^{-2 \lvl / 3} \brb{|\actions_{\lvl-1\to\lvl}| \ln |\expertree_\lvl|}^{1/3}
    $
    satisfies
        \begingroup
\setlength{\abovedisplayskip}{4pt} \setlength{\belowdisplayskip}{4pt}
    \[
        \reg{T}{2} \leq 32 \prtcst \lrb{ \sum_{\lvl =1}^{\nlvls} 2^{-2 \lvl / 3} \Brb{|\actions_{\lvl-1\to\lvl}| \ln |\expertree_\lvl| }^{1/3} }^{3/2} \sqrt{T} + \apprxcst 2^{-\nlvls} T \,,
    \]
    \endgroup
    where $|\actions_{\lvl-1\to\lvl}| := \sum_{a \in \actions_{\lvl-1}} |\suc_\lvl(a)|$. 
\end{theorem}
This provides a generic regret bound for any hierarchical tree satisfying Assumption~\ref{assump:function-tree}. The bound scales with level-wise quantities $|\actions_{\lvl-1\to\lvl}|$ and $|\expertree_\lvl|$ that will be made explicit in the next section to provide concrete rates. 
For now, observe that a naive application of the non-contextual algorithm of \cite{eldowa26a} would yield via their Corollary 4 a similar-looking bound with $\lrb{|\actions_{\lvl-1\to\lvl}| \ln |\expertree_\lvl| }^{1/3}$ replaced by $|\expertree_\lvl|^{1/3}$.
This improvement is thanks to \Cref{alg:2-point-feedback-experts} exploiting the problem structure; particularly, that information can be shared between policies such that in the end, what matters is the number of possible action pairs $|\actions_{\lvl-1\to\lvl}|$ instead of the number of policies.

A reminiscent result was obtained by \citet{zenati2023efficient} who considered an experts problem over a nested action set (extending the nested bandits setting of \citealt{martin2022nested}), and used an EXP4-type algorithm.
Their setting requires a stronger semi-bandit feedback, in which the learner observes $\nlvls$ intermediate losses induced by a given hierarchical decomposition of the played action. 
Nonetheless, in our notation, their regret upper bound (Theorem~5.5.1) would look like
\[
    \textstyle \sum_{\lvl = 1}^{\nlvls} 2^{-\lvl} \sqrt{|\actions_{\lvl-1\to\lvl}| T \ln |\expertree_\nlvls|} \geq 
    \frac{1}{\sqrt{L}}  \Big( \sum_{\lvl = 1}^{\nlvls} 2^{-2\lvl/3} \big({|\actions_{\lvl-1\to\lvl}| \ln |\expertree_\nlvls|}\big)^{1/3} \Big)^{3/2} \sqrt{T} \,.
\]
The additional multiplicative factor of order at most $\sqrt{\nlvls}$ that we incur is due to the weaker feedback we are considering (two points vs $\nlvls$ observations per round).
Crucially however, similar to what a more direct application of EXP4 would bring, this bound depends only on the logarithm of the total number of functions $\ln |\expertree_\nlvls|$; that is, it does not feature the refined dependence we have on $(\ln |\expertree_\lvl|)_{\lvl \leq \nlvls}$.
As an effect, one would not be able to achieve $\sqrt{T}$ regret using such a bound in the examples considered in the next sections (where $\ln |\expertree_\nlvls|$ is polynomial in $T$), which highlights the importance of relying on the policy tree and the nested regularizer.

\section{A Concrete Rate}
\label{sec:concrete_rate}
In this section, we provide a more explicit construction of the tree $\cT(\expertree_{0:\nlvls})$, which will be designed to satisfy \Cref{assump:function-tree} while keeping $\expertree_\lvl$ and $\actions_\lvl$ as small as possible.
As such, the cardinalities of these sets will be expressed in terms of fundamental complexity (or size) properties of $\actions$ and $\experts$.
Before proceeding, we recall some standard notions that we will be using next, all with respect to the metric $\adist$ on $\actions$ and the metric $\distfunc$ on policies from $\contexts$ to $\actions$.

\subsection{Complexity notions}

For $a \in \actions$ and $\epsilon > 0$, let $\ball(a;\epsilon) \coloneqq \bcb{ b \in \actions \mid \adist(a,b) \leq \epsilon }$ denote the $\epsilon$-ball centered at $a$.

\begin{definition}[Covers and packings]\label{def:covering}
    Let $A \subseteq \actions$ and $\epsilon > 0$.
    \begin{itemize}[nosep]
        \item An \emph{$\epsilon$-cover} of $A$ is a set $S \subseteq \actions$ such that $A \subseteq \bigcup_{a \in S} \ball(a;\epsilon)$, i.e., every point of $A$ is within distance $\epsilon$ of some point of $S$. The \emph{covering number} $\covering(A;\epsilon)$ is the cardinality of the smallest $\epsilon$-cover of $A$.
        \item An \emph{$\epsilon$-packing} of $A$ is a subset of $A$ whose distinct members $a \neq b$ all satisfy $\adist(a,b) > \epsilon$. The \emph{packing number} $\packing(A;\epsilon)$ is the cardinality of the largest $\epsilon$-packing of $A$.
    \end{itemize}
\end{definition}
Covers and packings are equivalent up to a factor $2$ in the scale:
$
    \packing(A;2\epsilon) \leq \covering(A;\epsilon) \leq \packing(A;\epsilon) \,.
$
The right inequality holds because a maximal $\epsilon$-packing is automatically an $\epsilon$-cover (a point of $A$ more than $\epsilon$ away from all its members could be added to it).
The left one holds because a ball $\ball(a;\epsilon)$ centered at a member of an $\epsilon$-cover cannot contain two members of a $2\epsilon$-packing, as they would be at distance at most $2\epsilon$ from each other.
We use maximal packings to construct the action sets $\actions_\lvl$, since they are both covers and well-separated (which we will require), and covering numbers to measure complexity.
For the policy classes, which are significantly larger, the relevant notion is the logarithm of the covering number in $(\functionspace,\distfunc)$, where $\functionspace$ is the set of all functions from $\contexts$ to $\actions$.

\begin{definition}[Metric entropy]
\label{def:metric_entropy}
    For $\functions \subseteq \functionspace$ and $\epsilon>0$, let $\covering(\functions;\epsilon)$ denote the covering number of $\functions$ in the metric space $(\functionspace, \distfunc)$.
    The {metric entropy} of $\functions$ at scale $\epsilon$ is $\metentropy(\functions;\epsilon) \coloneqq \ln \covering(\functions;\epsilon)$.
\end{definition}

\subsection{Design of the hierarchical tree and concrete rates}
With these notions in mind, we give now a general recipe for constructing the tree $\cT(\expertree_{0:\nlvls})$.
For the following, define $\extracst \coloneqq \sup_{a,b\in \actions \colon a \neq b} \adist(a,b)$, which is the diameter of $\actions$ \wrt $\adist$.
Firstly, fix a sequence $(\actions_\lvl)_{\lvl = 0}^\nlvls$ of subsets of $\actions$ such that $\actions_\lvl$ is a maximal  $\extracst 2^{-\lvl}$-packing of $\actions$. %
This implies that $\actions_\lvl$ is a $\extracst 2^{-\lvl}$-cover of $\actions$, and that $|\actions_\lvl| \leq \covering(\actions; \extracst 2^{-\lvl-1})$.
Next, we will choose a sequence of sets of functions $(\functions_\lvl)_{\lvl = 0}^\nlvls$ such that $\functions_\lvl$ is a $\extracst 2^{-\lvl+1}$-cover of $\experts$, $|\functions_\lvl| \leq \covering(\experts; \extracst 2^{-\lvl})$, and $\image(\func) \subseteq \actions_\lvl$ for all $\func \in \functions_\lvl$.
In particular, these sets can be constructed as follows.

Let $\tilde{\functions}_\lvl \subseteq \functionspace$ be a $\extracst 2^{-\lvl}$-cover of $\experts$ satisfying $\babs{\tilde{\functions}_\lvl} = \covering(\experts; \extracst 2^{-\lvl})$.
For any $\func \in \functionspace$, let $\disc_\lvl(\func) \in \functionspace$ be a discretization of $\func$ onto $\actions_\lvl$; that is,
$
    \disc_\lvl(\func)(x) \in \argmin_{a \in \actions_\lvl} \adist(a,\func(x))  
$
for all $x \in \contexts$.
Hence, $\distfunc\brb{\func,\disc_\lvl(\func)} \leq \extracst 2^{-\lvl}$.
So, we can choose $\functions_\lvl = \bcb{ \disc_\lvl(\func) \mid \func \in  \tilde{\functions}_\lvl}$, which is easily seen now (via the triangle inequality) to be a $\extracst 2^{-\lvl+1}$-cover of $\experts$ as required.
Note that since $\tilde{\functions}_\lvl$ is an optimal $\extracst 2^{-\lvl}$-cover of $\experts$, it must be that for every $\func \in \tilde{\functions}_\lvl$, there exists some $\expert \in \experts$ such that $\distfunc(\expert, \func) \leq \extracst 2^{-\lvl}$.
The same then holds for $\functions_\lvl$ at distance $\extracst 2^{-\lvl+1}$.

We will use $\functions_\lvl$ to populate $\expertree_\lvl$, the nodes of the tree at level $\lvl$.
To conclude the construction of the tree $\cT(\expertree_{0:\nlvls})$, we need to specify the parenthood relation.
Here, this can be any function such that for any $\func \in \functions_{\lvl+1}$, there exists some $\expert \in \experts$ satisfying $\distfunc(\expert, \func) \leq \extracst 2^{-\lvl}$ and $\distfunc(\expert, \parent(\func)) \leq \extracst 2^{-\lvl+1}$.
This is possible since, as remarked before, each member of $\functions_{\lvl+1}$ can be mapped to some $\expert \in \experts$ that is no more than $ \extracst 2^{-\lvl}$ far, and
$\functions_\lvl$ is a $\extracst 2^{-\lvl+1}$-cover of $\experts$.
As a consequence, for any $\func \in \functions_{\lvl+1}$, it holds via the triangle inequality that $\distfunc(\func, \parent(\func)) \leq 3 \extracst 2^{-\lvl}$.
We see now that our new tree satisfies \Cref{assump:function-tree} with $\apprxcst = 2 \extracst$ and $\prtcst = 3 \extracst$.
Then, building on \Cref{thm:generic-thm}, we establish the following guarantee for \Cref{alg:2-point-feedback-experts}. (The proof can be found in \Cref{app:concrete}.)

\begin{restatable}{recor}{thmconcrete}
    \label{thm:concrete}
    Fix $\covcst,\funcst >0$
    and take 
    \begingroup
\setlength{\abovedisplayskip}{4pt} \setlength{\belowdisplayskip}{4pt}
    \begin{align*}
   &  \covdim  = \inf \bcb{ r \geq 0 \mid \covering(\actions;\epsilon) \leq \covcst \brb{\extracst / \epsilon}^{r} \:\forall\epsilon \in (0,\extracst] } \qquad  \\
     \text{ and, \qquad} & \fundim  = \inf \bcb{ r \geq 0 \mid \metentropy(\experts; \epsilon) \leq \funcst \brb{\extracst / \epsilon}^{r} \:\forall\epsilon \in (0,\extracst] } \,.
    \end{align*}
\endgroup
     Assume that $\log_2(T) \geq (\covdim + \fundim) \lor 2$.
    Then, tuned as in  \Cref{thm:generic-thm} with $\apprxcst = 2 \extracst$ and $\prtcst = 3 \extracst$, \Cref{alg:2-point-feedback-experts} with $\cT(\expertree_{0:\nlvls})$ as described above and $\nlvls =  \bfl{{\log_2(T)}/{((\covdim + \fundim) \lor 2 )}}$ satisfies
    \begin{align*}
        \reg{T}{2} \leq 4 \extracst \lrb{ 24  \sqrt{\covcst \funcst} \, 4^{\dbldim} + 1}  \logfactor(T,\covdim,\fundim)^{\frac{3}{2}} \, T^{1-\frac{1}{(\covdim + \fundim) \lor 2 }} \,, 
    \end{align*}
    where $\dbldim \coloneqq \inf \bcb{ d \geq 0 \mid \covering\brb{\ball(a; \epsilon); \epsilon/2} \leq 2^{d} \: \forall  a \in \actions, \epsilon > 0}$ is the doubling dimension of $(\actions,\adist)$, and $ \logfactor(T,\covdim,\fundim) \coloneqq \babs{1-2^{-(\covdim + \fundim-2)/{3}}}^{-1} \land {\log_2(T)}/{ 2 } \,.$
\end{restatable}
Essentially, the rate is $\sqrt{T}$ when $\covdim + \fundim \leq 2$ and $T^{1-{1}/(\covdim + \fundim)}$ otherwise, up to a $(\log_2(T))^{3/2}$ factor that appears when $\covdim + \fundim$ is around $2$. 
Here, $\covdim$ is the covering dimension of $\actions$ with multiplicity $\covcst$, and $\fundim$ is the metric entropy exponent of $\experts$ with multiplicity $\funcst$.
The derivation mostly follows from \Cref{thm:generic-thm} and the cardinality bounds we enforced in the construction.
The crux is showing that $\suc_{\lvl}(a)$ is bounded in terms of the doubling constant uniformly over $\lvl \in [\nlvls]$ and $a \in \actions$.  
This holds thanks to the proximity of a function and its parent, combined with the fact that, by construction, the image of a function at any level is a packing. 

We now instantiate the bound of \Cref{thm:concrete} for some standard policy classes.
We assume here that $\actions$ is a compact subset of $\R^{\adim}$ and that $\adist$ is the Euclidean distance.
Using a uniform grid discretization, one can show that $\covering(\actions;\epsilon) \lesssim \epsilon^{-\adim}$
up to a constant depending on $\adim$ and the diameter of $\actions$.
Moreover, the doubling dimension $\dbldim$ in this case is known to be $\cO(\adim)$ \citep{gupta2003}.
As for the contexts, we take $\contexts$ as a compact subset of $\R^\cdim$.

\paragraph{Lipschitz Functions}
Take the policy space to be the set of $1$-Lipschitz functions mapping contexts to actions; so that
$\|\expert(x) - \expert(x')\| \leq  \|x-x'\|$ for $\expert \in \experts$ and $x,x' \in \contexts$.
It can be shown that $\metentropy(\experts;\epsilon) \lesssim \epsilon^{-\cdim}$ with a constant depending on $\adim$, $\cdim$, and the diameters of $\actions$ and $\contexts$;
an explicit construction is provided in \Cref{sec:efficient}.
\Cref{thm:concrete} then implies that $\reg{T}{2}$ is $\cO(\ln(T)^{3/2}  \sqrt{T})$ when $\adim=\cdim=1$ (essentially matching the full-information rate) and $\smash{\cO(T^{1-{1}/(\adim + \cdim)})}$ otherwise.

\paragraph{H\"older-Smooth Functions}
Fix constants $\derbound, \holder, \holdercst > 0$ and let $\ders$ be the largest integer strictly smaller than $\holder$.
Take $\actions = [-\derbound,\derbound]$ and $\contexts = [0,1]^\cdim$.
The policy set $\experts$ now will consist of functions $\expert \colon \contexts \rightarrow \R$ such that for every $\dervec = (\derelem_1,\dots,\derelem_\cdim)$ where $\derelem_i$ is an integer $\geq 0$ and $\dersum \coloneqq \sum_{i=1}^\cdim \derelem_i \leq \ders$, 
$
    \sup_{x \in \contexts} |D^\dervec \expert(x)| \leq \derbound 
$,
where
$
    D^\dervec  \coloneqq \frac{\partial^{\dersum} }{\partial x_1^{\derelem_1} \dots \partial x_\cdim^{\derelem_\cdim}} \,;
$
moreover, %
\begingroup
\setlength{\abovedisplayskip}{5pt} \setlength{\belowdisplayskip}{4pt}
\[
    \sup_{\dersum = \ders} \sup_{x,y \in \contexts} \abs{ D^\dervec \expert(x) - D^\dervec \expert(y) } \leq \holdercst \norm{x-y}^{\holder - \ders} \,.
\]
\endgroup
Note that Lipschitz functions are recovered when $\holder = 1$.
It holds here that $\metentropy(\experts;\epsilon) \lesssim \epsilon^{-\cdim/\holder}$ with a constant depending on $\cdim$, $\beta$, $\derbound$, and $\holdercst$;
see Theorem 2.7.1 in \cite{van1996} or Theorem 2 in
\cite{lorentz1962}.
So, we can take $\fundim = \cdim / \holder$ and $\covdim =1$, in which case
\Cref{thm:concrete} gives that $\reg{T}{2}$ is $\cO(\sqrt{T})$ when $\holder > \cdim$, $\cO(\ln(T)^{3/2}  \sqrt{T})$ when $\holder = \cdim$, and $\cO\brb{T^{{\cdim}/(\cdim+ \holder)}}$ otherwise.
In the first two cases, we match the full-information rate up to log factors.

\section{An Efficient Implementation for Lipschitz Policies}\label{sec:efficient}
In most interesting cases, the tree built in \Cref{sec:concrete_rate} is enormous and hence \Cref{alg:2-point-feedback-experts} would be intractable to implement.
In this section, we focus on Lipschitz policy spaces and show that we can mimic the behaviour of \Cref{alg:2-point-feedback-experts} in an efficient manner when $\adim$ is small.
This is possible by giving the function tree a structure that is separable in a certain sense, which allows us to devise more handy representations for objects acting on its constituent functions.
For a clear presentation we will consider a simple setup with $L_\infty$ geometry.
Concretely, we assume that $\contexts = [0,1]^\cdim$, $\actions = [0,1]^\adim$, and $\adist(a,b) = \losslip \norm{a - b}_\infty$. 
This implies that
$\losses = \{ \loss \colon \actions \rightarrow \R \mid |\loss(a) - \loss(b)| \leq \losslip \norm{a - b}_\infty \:\forall a,b \in \actions \} $, 
i.e., the set of $\losslip$-Lipschitz functions from $[0,1]^\adim$ to $\R$ according to the $L_\infty$ metric. 
Lastly, assume that $\experts$ is the set of $1$-Lipschitz (in $L_\infty$) functions mapping contexts to actions:
$
    \experts \coloneqq \bcb{ \expert \colon \contexts \rightarrow \actions \mid \|\expert(x) - \expert(x')\|_\infty \leq  \|x-x'\|_\infty \:\forall \, x,x' \in \contexts }\,.
$

\subsection{Building a structured tree}\label{sec:efficient:tree}
We start by building a nested partitioning (or a hierarchical discretization) of the context space.
(Analogous approaches can be found, e.g., in \citealp{gaillard2015chaining}.)
For $\lvl \geq 0$, define $\vertexset_\lvl \coloneqq \{ (\vertex_\varlvl)_{\varlvl\leq\lvl} \mid \vertex_0 = \ones_\cdim \text{ and } \vertex_\varlvl \in \{-1,1\}^\cdim\}$, which satisfies $|\vertexset_\lvl| = 2^{\lvl \cdim }$. 
A member of this set is essentially an address of a cell (subset) of $\contexts$ in the partitioning associated with level $\lvl$.
In particular, a sequence $\vertices = (\vertex_\varlvl)_{\varlvl\leq\lvl} \in \vertexset_\lvl$ can be used to identify a context $\ctx(\vertices) \coloneqq \sum_{\varlvl \leq \lvl } 2^{-\varlvl-1} \vertex_\varlvl \in \contexts$ and a cell which is the $L_\infty$ ball of radius $2^{-\lvl-1}$ centered around $\ctx(\vertices)$.\footnote{Precisely, to obtain non-overlapping boundaries, we define 
$
    \cell(\vertices) \coloneqq \bcb{ x \in \contexts \mid \forall i \in [\cdim]\,, \brb{\floor{2^{\lvl} x_i} + 1 } \land 2^\lvl = \floor{2^{\lvl}  \ctx(\vertices)_i} + 1
    } 
$.}
The cells $\{\cell(\vertices)\}_{\vertices \in \vertexset_\lvl}$ form a uniform grid of $\contexts$ with $2^{-\lvl}$ steps, and it holds that $\max_{x \in \cell(\vertices)} \norm{x -\ctx(\vertices)}_\infty \leq 2^{-\lvl-1}$.

Fix $\vertices = (\vertex_{\varlvl})_{\varlvl \leq \lvl} \in \vertexset_\lvl$.
The sub-sequence of $\vertices$ up to level $\lvl'$ will be denoted by $\vertices_{\upto \lvl'}$, i.e., $\vertices_{\upto \lvl'} = (\vertex_{\varlvl})_{\varlvl \leq \lvl'}$.
We will use $\rootvertex$ to denote the trivial sequence $(\ones_\cdim)$, so that 
$\vertices_{\upto 0} = \rootvertex$.
The set of children of $\vertices$ is given by 
$\children(\vertices) = \{ \vertices' \in \vertexset_{\lvl+1} \mid \vertices'_{\upto \lvl} = \vertices\}$, 
and
its parent is $\parent(\vertices) 
= \vertices_{\upto \lvl-1} \in \vertexset_{\lvl-1}$. 
It holds then that $\norm{\ctx(\vertices') -\ctx(\vertices)}_\infty \leq 2^{-\lvl-2}$ for all $\vertices' \in \children(\vertices)$.
A crucial property is that  $ \cell(\vertices) \subset \cell(\parent(\vertices))$; meaning that the partitions are nested across levels.
For $x \in \contexts$, let $\seq(x) = (\seq_\varlvl(x))_{\varlvl\leq\nlvls}$ denote the unique node in $\vertexset_\nlvls$ to whose cell $x$ belongs, and let $\seq_{\upto\lvl}(x) = (\seq_\varlvl(x))_{\varlvl\leq\lvl}$ be its ancestor at level $\lvl$. 

For convenience, we will give ourselves the freedom of selecting actions within the extended set $\actions^+ = [-1,2]^\adim$ with the understanding that $\loss(a) = \loss(\clip(a))$, where $\clip(a)_i = 1 \land (0 \lor a_i)$.
These extended loss functions are still $\losslip$-Lipschitz since \[
|\loss(a) - \loss(b)| = |\loss(\clip(a)) - \loss(\clip(b))| \leq \losslip \norm{\clip(a) - \clip(b)}_\infty \leq \losslip \norm{a - b}_\infty \,.
\]
Moving over to the function tree, we will rely on the following structured representation:
\[
    \expertree_\lvl \coloneqq \Bcb{ \config = (\prof_\vertices)_{\vertices\in \vertexset_{0:\lvl}} \mid \prof_{\rootvertex} = 2^{-1} \ones_\adim \text{ \& } \prof_\vertices \in   \prof_{\parent(\vertices)}  + 2^{-\varlvl-1} \cdot \{-3,-1,1,3\}^\adim \, \forall \varlvl \leq \lvl,\vertices \in \vertexset_{\varlvl}
    } \,.
\]
Each member of this set is an assignment of actions, one per context node, such that the actions assigned to a context node and its child satisfy a certain proximity relationship.
The choice of taking steps in $2^{-\varlvl-1} \cdot \{-3,-1,1,3\}^\adim$ instead of the seemingly more natural $2^{-\varlvl-1} \cdot \{-1,1\}^\adim$ is made to ensure that $\expertree_\nlvls$ covers the policy space, as required by \Cref{assump:function-tree}.
Each node $\config \in \expertree_\lvl$ encodes the step-wise constant function defined for all $x \in \contexts$ by
\begingroup
\setlength{\abovedisplayskip}{5pt} \setlength{\belowdisplayskip}{5pt}
\[
    \func_{\config}(x) := \summ_{\vertices \in \vertexset_\lvl} \indicator{x \in \cell(\vertices)} \prof_\vertices\,,
\]
\endgroup
which gives us that $\suc_{\lvl}(a) = \{ a + 2^{-\lvl-1}\diff \mid \diff \in \{-3,-1,1,3\}^\adim \}$ for $a \in \actions_{\lvl-1}$.
One can then take $\actions_0 = \{ 2^{-1} \ones_\adim \}$ and 
$\actions_\lvl = \bigcup_{a \in \actions_{\lvl-1}} \suc_{\lvl}(a) $, where the latter can be written explicitly as $\actions_\lvl = \bcb{2^{-\lvl} i - 2^{-\lvl-1} \mid i \in \{-2^\lvl+2, \dots, 2^{\lvl+1}-1\}}^\adim$, which always lies inside $\actions^+$.
Observe that $|\actions_\lvl| = (3 \cdot 2^{\lvl} - 2)^\adim$ and $|\suc_{\lvl}(a)| = 4^\adim$.
An easy count also yields that $\ln |\expertree_\lvl| 
\leq \ln(4) \adim 2^{(\lvl+1) \cdim} $.

The usual relations between the nodes in different levels are simple and natural to specify here.
The parent of $\config \in \expertree_\lvl$ is $\parent(\config) = (\prof_\vertices)_{\vertices\in \vertexset_{0:\lvl-1}} \in \expertree_{\lvl-1}$, and its children set is 
\[\children(\config) = \bcb{ \config' = (\prof'_\vertices)_{\vertices\in \vertexset_{0:\lvl+1}} \mid \prof'_{\vertices} = \prof_{\vertices} \text{ for } \vertices\in \vertexset_{0:\lvl} \text{ and } \prof'_{\vertices} \in \suc_{\lvl+1}(\prof_{\parent(\vertices)}) \:\text{ for } \vertices\in \vertexset_{\lvl+1} } \,.
\]
In a similar manner, for any $\lvl \in \{0,\dots,\nlvls\}$ and $\config \in \bigcup_{\varlvl \geq \lvl} \expertree_{\varlvl}$, 
the ancestor of $\config$ at level $\lvl$ is given by $\ancfun_{\lvl}(\config) = (\prof_\vertices)_{\vertices\in \vertexset_{0:\lvl}} \in \expertree_\lvl$.
This tree, as shown in \Cref{app:efficient:tree-props}, satisfies \Cref{assump:function-tree} with $\apprxcst = \losslip$ and $\prtcst = 3 \cdot 2^{-2} \losslip$. 
With all of this in hand, it is easy to show via \Cref{thm:generic-thm} that \Cref{alg:2-point-feedback-experts} enjoys a regret rate of $T^{1-1/(\adim+\cdim)}$ on this tree, up to log factors, if we set $\nlvls \approx \frac{1}{\adim+\cdim} \log_2 T$. 
The issue now is of course the computational complexity of the algorithm seeing as it requires handling objects of size $|\expertree_\nlvls|$, which is of order $\exp(T^{{\cdim}/(\adim+\cdim)})$.
We now present a more efficient implementation of essentially the same algorithm, leveraging the structure of this tree.

\subsection{An efficient algorithm and its regret bound}\label{sec:efficient:alg}
The loss estimates of \Cref{alg:2-point-feedback-experts} can be written here as
\[
    \hat{\expertloss}_t(\config) = \frac{\indicator{  \bee_{t,1} = \prof_{\activenode_{t}} \,,\, \bee_{t,2}  = \prof_{\parent(\activenode_t)} }}{\delta_{\lvl_t} \pr_t[ \bee_{t,1} = \prof_{\activenode_{t}} \,,\, \bee_{t,2}  = \prof_{\parent(\activenode_t)}    \mid \lvl_t]} \brb{\loss_t(A_{t,1}) - \loss_t(A_{t,2}) + b_{\lvl_t}}
\]
where $\activenode_{t} = \seq_{\upto\lvl_t}(x_t)$.
It is clear that this function depends on its input $\config$ only through the actions it assigns to the child-parent nodes $\activenode_{t}$ and $\parent(\activenode_{t})$.
Now, for each context node $\vertices \in \vertexset_\lvl$, we can define a function $\closs_{t,\vertices} \colon \actions_{\lvl} \times \actions_{\lvl-1} \rightarrow \R$ as
\[
    \closs_{t,\vertices}(b,a) \coloneqq \frac{\indicator{\vertices = \activenode_{t} \,,\,  \bee_{t,1} = b \,,\, \bee_{t,2}  = a }}{\delta_{\lvl_t} \pr_t[ \bee_{t,1} = b \,,\, \bee_{t,2}  = a    \mid \lvl_t]} \brb{\loss_t(A_{t,1}) - \loss_t(A_{t,2}) + b_{\lvl_t}} \,,
\]
which clearly gives that
$
    \hat{\expertloss}_t(\config) = \sum_{\vertices \in \vertexset_{1:\nlvls}} \closs_{t,\vertices}(\prof_{\vertices},\prof_{\parent(\vertices)})
$.
Summing up to round $t$,
we have that 
$\sum_{s \leq t} \genloss_s(\config) = \sum_{\vertices \in  \vertexset_{1:\nlvls}} \closses_{t,\vertices}(\prof_\vertices,\prof_{\parent(\vertices)})$,
where $\closses_{t,\vertices}(b,a) \coloneqq \sum_{s \leq t} \closs_{s,\vertices}(b,a)$.
Thanks to this separable representation of the loss function, we show in \Cref{app:efficient:dec} that using the regularizer chosen in \Cref{thm:generic-thm} gives a closed form decomposition of $p_{t+1}$:
\[
    p_{t+1}(\config) = \prod_{\vertices \in \vertexset_{1:\nlvls}} \cond_{t+1,\vertices}(\prof_{\vertices} \mid \prof_{\parent(\vertices)}) \text{ with } \cond_{t+1,\vertices}(b \mid a) \propto \exp\Brb{- \etas^{-1}_{\lvl} \brb{\closses_{t,\vertices}(b,a) + \omg_{t,\vertices}(b)}}
\]
for $\vertices \in \vertexset_\lvl$, $a \in \actions_{\lvl-1}$, and $b \in \suc_{\lvl}(a)$.
Here, $\etas_\lvl := \sum_{\varlvl \geq \lvl} {1}/{\eta_{\varlvl}}$ and $\omg_{t,\vertices}$ is a recursive function (see \eqref{def:efficient:omega} below) defined in terms of $\omg_{t,\vertices'}$ and $\closses_{t,\vertices'}$ at children nodes $\vertices'$.

In words, $\cond_{t+1,\vertices}(b \mid a)$ is the probability that a policy $\config$ sampled from $p_{t+1}$ plays $b$ at $\vertices$ given that it plays $a$ at $\parent(\vertices)$.
Storing these distributions (specifically, the functions $\closses_{t,\vertices}$ and $\omg_{t,\vertices}$) is drastically cheaper than storing $p$ itself.
Further, most of these functions will be sparsely updated.
This factorization also simplifies the process of sampling, especially when we are only looking to evaluate the sampled function (and its ancestors) on a specific context.
In particular, fix $x \in \contexts$ and let $\vertices = \seq(x)$.
Then, if $\bm{\vartheta} \in \expertree_\nlvls$ is a random variable distributed according to $p_{t}$, a sample of the sequence $(\bm{\vartheta}_{\vertices_{:\lvl}})_{\lvl\leq \nlvls}$ can be obtained by sequentially sampling a path $(\action_\lvl)_{\lvl\leq \nlvls}$ with $\action_0 = 2^{-1} \ones_\adim$ and $\action_\lvl \sim \cond_{t,\vertices_{:\lvl}}(\cdot \mid \action_{\lvl-1})$.
\Cref{alg:efficient} employs these ideas to emulate \Cref{alg:2-point-feedback-experts} cheaply and still enjoy a regret bound of the same order as that presented in \Cref{sec:concrete_rate}: (see \Cref{app:efficient:regret} for a proof)

\LinesNumbered
\begin{algorithm}[!t] 
 \KwIn{tree $\cT(\expertree_{0:\nlvls})$, constants $(b_\lvl)_{\lvl=1}^\nlvls$, learning rates $(\eta_\lvl)_{\lvl=1}^\nlvls$, distribution $\delta \in \Delta_{\nlvls}$}
 
 \KwInit{$\closses_{0,\cnode}(b,a) = 0$ for $\lvl \in [\nlvls], \cnode \in \cnodes_\lvl$, $a \in \actions_{\lvl-1}$, and $b \in \suc_{\lvl}(a)$}

 compute $\omg_{0,\cnode}(a)$ for $\lvl \in \{0,\dots,\nlvls\}, \cnode \in \cnodes_\lvl$, and $a \in \actions_{\lvl}$ via \eqref{def:efficient:omega}

 \For{$t \geq 1$}{

    observe $x_t$ and set $\vertices = \seq(x_t)$

    set $\action_{t,0} = 2^{-1} \ones_\adim $

    \lFor{$\lvl =1,\dots,\nlvls$}
    {
        sample $\action_{t,\lvl}$ from $\cond_{t,\vertices_{\upto\lvl}}(\cdot \mid \action_{t,\lvl-1})$
    }

    sample $\lvl_t \sim \delta$ and set $\action'_{t,\lvl} = \action_{t,\lvl}$ for $\lvl = 0, \dots,\lvl_t-1$

    \lFor{$\lvl =\lvl_t,\dots,\nlvls$}
    {
        sample $\action'_{t,\lvl}$ from $\cond_{t,\vertices_{\upto\lvl}}(\cdot \mid \action'_{t,\lvl-1})$
    }

    play $A_{t,1} = \action_{t,\nlvls}$ and $A_{t,2} = \action'_{t,\nlvls}$; set $\bee_{t,1} = \action_{t,\lvl_t}$ and $\bee_{t,2} = \action_{t,\lvl_t-1}$

    set $\displaystyle \closses_{t,\activenode_t}(b,a) = \closses_{t-1,\activenode_{t}}(b,a) +  \indicator{\bee_{t,1}=b, \bee_{t,2}=a}\frac{\loss_t(A_{t,1}) - \loss_t(A_{t,2}) + b_{\lvl_t}} {\delta_{\lvl_t} \pr_t[ \bee_{t,1} = b,\bee_{t,2} = a  \mid \lvl_t]}$, \\
    where $\activenode_{t} = \vertices_{\upto\lvl_t}$, $a \in \actions_{\lvl_{t-1}}$, and $b \in \suc_{\lvl_t}(a)$
    
    update $\omg_{t,\vertices'}(a)$ for $\lvl \in \{0,\dots,\nlvls\}, \vertices' \in \cnodes_\lvl$, and $a \in \actions_{\lvl}$ via \eqref{def:efficient:omega}
 }
 \caption{Factorized Policy-Tree Exponential Weights with Two-Point Feedback} \label{alg:efficient}
\end{algorithm}

\begin{restatable}{rethm}{thmefficient}\label{thm:efficient}
        Assuming $\log_2(T) \geq 4(\adim +\cdim)$, 
        \Cref{alg:efficient} with $b_\lvl = 3 \losslip 2^{-\lvl+1}$, $\delta_\lvl \propto 2^{\lvl(\cdim+\adim-2)/3}$, $
        \eta_{\lvl} = \frac{2^{\lvl}}{12 \losslip}\sqrt{\frac {\ln(4) \adim \delta_\lvl  2^{(\lvl+1)(\cdim-\adim)}}{ 3^\adim T } }
        $, and $\nlvls =  \bfl{\frac{\log_2(T)}{\cdim + \adim } - 3}$
    satisfies 
    \begin{equation*}
        \reg{T}{2} \leq 16 \losslip \cdot
        \begin{cases}
            \brb{1 + 12 \sqrt{\ln(4)} (\log_2(T)/2)^{3/2} } \sqrt{T} &\text{if } \adim=\cdim=1 \\
            \brb{1+ 132 \sqrt{\ln(4) \adim} } T^{1-\frac{1}{\adim + \cdim}} &\text{otherwise}
        \end{cases}\,.
    \end{equation*}  
\end{restatable}

\paragraph{On the computational complexity}\label{sec:efficent:complexity}
We now argue that \Cref{alg:efficient} is efficient: it has a space complexity of $\cO(T \ln T)$ and a per-round time complexity of $\cO(\ln T)$.
Here we assume that $\adim$ is small and treat it as constant.
Observe that $\nlvls \lesssim \ln T$, $|\vertexset_\nlvls| \lesssim T^{\cdim/(\adim+\cdim)}$, $|\actions_{\nlvls}| \lesssim T^{\adim/(\adim+\cdim)}$, and the size of the successor set of any action is $\cO(1)$.
Hence, storing the functions $\omg_{t,\vertices}$ and $\closses_{t,\vertices}$ for all $\vertices \in \vertexset_{0:\nlvls}$ has a complexity of at most $\cO(T \ln T)$.
Having stored these functions, sampling the two paths $(\action_{t,\lvl})_{\lvl\leq\nlvls}$ and $(\action'_{t,\lvl})_{\lvl\leq\nlvls}$
requires $\cO(\ln T)$ computations since the support of $\cond_{t,\vertices}(\cdot \mid a)$ is the successor set of $a$.
Another burdensome step is updating the functions $\omg_{t,\vertices}$.
For $\lvl \in \{0,\dots,\nlvls-1\}$, $\vertices \in \vertexset_{\lvl}$, and $a \in \actions_{\lvl}$, $\omg_{t, \vertices}(a)$ is defined as (see \Cref{lem:efficient:exp-profiles-closed-form} in \Cref{app:efficient:dec})
\begingroup
\setlength{\abovedisplayskip}{2pt} \setlength{\belowdisplayskip}{4pt}
    \begin{align} \label{def:efficient:omega}
        \omg_{t, \vertices}(a) \coloneqq \sum_{\vertices' \in \children(\vertices)} -\etas_{\lvl+1} \ln \lrb{\sum_{b \in \suc_{\lvl+1}(a)} \exp\Brb{-\etas^{-1}_{\lvl+1} \brb{\closses_{t,\vertices'}(b,a) + \omg_{t,\vertices'}(b)}}} \,,
    \end{align}
    \endgroup
    while for $\vertices \in \vertexset_{\nlvls}$ and $a \in \actions_{\nlvls}$, set $\omg_{t,\vertices}(a) \coloneqq 0$. We discuss next the complexity of these updates.
    
    Recall that we denote by $\seq(x_t)$ the sequence of context nodes (at each level of the tree) to which $x_t$ belongs.
    Now $\closses_{t,\vertices}$ will only be updated (i.e., incremented compared to its value at the previous round) for $\vertices$ lying in said path, and hence the same holds for $\omg_{t,\vertices}$.
    In fact, only $\displaystyle \closses_{t,\activenode_t}(\bee_{t,1},\bee_{t,2})$ is updated at round $t$, where $\activenode_t = \seq_{\upto \lvl_t}(x_t)$.
    This implies via \eqref{def:efficient:omega} that $\closses_{t,\vertices}$ and $\omg_{t,\vertices}$ will remain unchanged for all nodes in $(\seq_{\upto \lvl}(x_t))_{\lvl\geq\lvl_t}$.
    It is $\omega_{t,\parent(\activenode_t)}$ that is first updated, and only at the action $\bee_{t,2}$.
    This change must then be propagated upwards to $\omg_{t,\vertices}$ for all nodes $\vertices$ in $(\seq_{\upto \lvl}(x_t))_{\lvl\leq\lvl_t-2}$.
    Inspecting \eqref{def:efficient:omega} again, we can see that at any of these nodes, say node $\vertices$ at level $\lvl$, $\omg_{t,\vertices}$ needs to be updated only at actions $a \in \actions_\lvl$ from which it is possible to construct a downwards path $(a,\dots,\bee_{t,2})$ to $\lvl_t-1$ with each action belonging to the successor set of the previous.
    In \Cref{app:efficient:pre}, we call this set of actions $\pre_{\lvl}(\bee_{t,2})$ and show that, luckily, $|\pre_{\lvl}(\bee_{t,2})| \leq 3^\adim$ uniformly at all levels.
    This implies that we only need to carry out $\cO(\ln T )$ updates.

    Finally, given $\lvl \in [\nlvls]$, $a \in \actions_{\lvl-1}$, and $b \in \suc_{\lvl}(a)$; we need to be able to compute $\pr_t(\bee_{t,1} = b, \bee_{t,2} = a \mid \lvl_t = \lvl)$.
    Since $\bee_{t,1} = \action_{t,\lvl_t}$, $\bee_{t,2} = \action_{t,\lvl_t-1}$, and $\lvl_t$ is independent of $(\action_{t,\lvl})_{\lvl \leq \nlvls}$; it holds that $\pr_t(\bee_{t,1} = b, \bee_{t,2} = a \mid \lvl_t = \lvl) = \pr_t(\action_{t,\lvl} = b, \action_{t,\lvl-1} = a) = \pr_t(\action_{t,\lvl-1} = a) \cond_{t,\seq_{\upto \lvl}(x_t)}(b \mid a)$.
    On the other hand, 
    $\pr_t(\action_{t,\lvl-1} = a) = \sum_{a' \in \pre_{\lvl-2}(a)} \pr_t( \action_{t,\lvl-2} = a')\cond_{t,\seq_{\upto \lvl-1}(x_t)}(a \mid a')$.
    Iterating this upwards, we see that, in addition to $\cond_{t,\seq_{\upto \lvl}(x_t)}(b \mid a)$, we need to compute the conditional probabilities $\cond_{t,\seq_{\upto \varlvl}(x_t)}(c \mid c')$ for all pairs $c \in \pre_{\varlvl}(a),  c' \in \pre_{\varlvl-1}(a)$ for $\varlvl = \lvl-1,\dots,1$.
    Since, again, $\pre_{\varlvl}(\cdot)$ is uniformly bounded independently of $\varlvl$, we conclude that the complexity of this procedure is still $\cO(\ln T)$.
    All in all, we see that the per-round complexity of \Cref{alg:efficient} is $\cO(\ln T)$.

\section*{AI Disclosure}
Some elements of the efficient algorithm of \Cref{sec:efficient} were arrived at thanks to interactions with Claude (various models). 
Otherwise, AI use was limited to proofreading and proof verification.

\acks{The French government  partly funded this work under the management of Agence Nationale de la Recherche as part of the
“France 2030” program, reference ANR-23-IACL-0006 (MIAI Cluster).}

\bibliography{biblio}

\newpage

\appendix

\crefalias{section}{appendix} %
\crefalias{subsection}{appendix} %

\section{Analysis of Algorithm \ref{alg:2-point-feedback-experts}}\label{app:generic-analysis}
In this appendix, we present the analysis of \Cref{alg:2-point-feedback-experts} leading to the regret bound of \Cref{thm:generic-thm}.
\begin{lemma} 
\label{lem:same-marginals-independent-from-level}
    It holds that $\pr_t( \func_{t,1} = \cdot ) = \pr_t(\func_{t,2} = \cdot ) = p_t(\cdot)$, $\func_{t,1} \ind \lvl_t \mid \cF_{t-1}$, and $\func_{t,2} \ind \lvl_t \mid \cF_{t-1}$.
\end{lemma}
\begin{proof}
    Fix any $\func \in \expertree_\nlvls$.
    Clearly, $\pr_t(\varfunc_{t} = f) = p_t(\func)$.
    We now show that the same holds for $\func_{t,1}$.
    To start with, fix a level $\lvl \in [\nlvls]$ and observe that
    \begin{align*}
        \pr_t(\func_{t,1} = f \mid \lvl_t = \lvl) &= \sum_{a \in \actions_{\lvl}} \pr_t\brb{ \anc{\varfunc_{t}}{\lvl}(x_t) = a \mid \lvl_t = \lvl} \pr_t\brb{ \func_{t,1} = \func \mid \anc{\varfunc_{t}}{\lvl}(x_t) = a , \lvl_t = \lvl} \\
        &= \sum_{a \in \actions_{\lvl}} \pr_t\brb{ \anc{\varfunc_{t}}{\lvl}(x_t) = a} \pr_t\brb{ \func_{t,1} = \func \mid \anc{\varfunc_{t}}{\lvl}(x_t) = a , \lvl_t = \lvl} \\
        &= \sum_{a \in \actions_{\lvl}} \pr_t\brb{ \anc{\varfunc_{t}}{\lvl}(x_t) = a }  \frac{\indicator{\anc{\func}{\lvl}(x_t) = a} p_t(\func)}{\sum_{\func' \in \expertree_\nlvls} \indicator{\anc{\func'}{\lvl}(x_t) = a} p_t(\func')} \\
        &= p_t(\func) \frac{\pr_t\brb{ \anc{\varfunc_{t}}{\lvl}(x_t) = \anc{\func}{\lvl}(x_t) }}{\sum_{\func' \in \expertree_\nlvls} \indicator{\anc{\func'}{\lvl}(x_t) = \anc{\func}{\lvl}(x_t)} p_t(\func')} = p_t(\func) \,,
    \end{align*}
    where the second equality uses that $\varfunc_t$ and $\lvl_t$ are sampled independently; the third equality follows from the sampling rule for $\func_{t,1}$ in \Cref{alg:2-point-feedback-experts}; and the last equality holds since $\pr_t(\varfunc_{t}=\cdot) = p_t(\cdot)$.
    As the resulting expression does not depend on $\lvl$, it follows that $\func_{t,1} \ind \lvl_t \mid \cF_{t-1}$ and that
    \[
        \pr_t(\func_{t,1} = f) = \sum_{\lvl \in [\nlvls]} \delta_\lvl \pr_t(\func_{t,1} = f \mid \lvl_t = \lvl) = p_t(\func) \,.
    \]
    The same can be shown for $\func_{t,2}$ in a similar manner using $\anc{\cdot}{\lvl-1}$ instead of $\anc{\cdot}{\lvl}$.
    
\end{proof}

\begin{lemma}\label{lem:loss-diff-bound}
    It holds at every round that
    $
        \abs{ \loss_t(A_{t,1}) - \loss_t(A_{t,2}) } \leq \prtcst 2^{-\lvl_t + 3} \,.
    $
\end{lemma}
\begin{proof}
    Recall that $A_{t,1} = \func_{t,1}(x_t) $ and $ A_{t,2} = \func_{t,2}(x_t)$, and also that
    \[
        \anc{\func_{t,1}}{\lvl_t}(x_t) = \anc{\varfunc_{t}}{\lvl_t}(x_t)  \qqand \anc{\func_{t,2}}{\lvl_t-1}(x_t) = \anc{\varfunc_{t}}{\lvl_t-1}(x_t) \,.
    \]
    Hence, via the triangle inequality
    \begin{align*}
        &\adist\brb{A_{t,1},\anc{\varfunc_{t}}{\lvl_t-1}(x_t)} \\
        &\hspace{4em}= \adist\brb{\func_{t,1}(x_t),\anc{\varfunc_{t}}{\lvl_t-1}(x_t)} \\
        &\hspace{4em}= \adist\brb{\anc{\func_{t,1}}{\lvl_t}(x_t),\anc{\varfunc_{t}}{\lvl_t-1}(x_t)} + \sum_{\lvl = \lvl_t+1}^{\nlvls} \adist\brb{\anc{\func_{t,1}}{\lvl}(x_t),\anc{\func_{t,1}}{\lvl-1}(x_t)} \\
        &\hspace{4em}= \adist\brb{\anc{\varfunc_{t}}{\lvl_t}(x_t),\anc{\varfunc_{t}}{\lvl_t-1}(x_t)} + \sum_{\lvl = \lvl_t+1}^{\nlvls} \adist\brb{\anc{\func_{t,1}}{\lvl}(x_t),\anc{\func_{t,1}}{\lvl-1}(x_t)} \\
        &\hspace{4em}\leq \prtcst 2^{-\lvl_t+1} + \sum_{\lvl = \lvl_t+1}^{\nlvls} \prtcst 2^{-\lvl+1} \leq \prtcst 2^{-\lvl_t + 2} \,,
    \end{align*}
    where the first inequality uses the second part of \Cref{assump:function-tree} seeing as $\anc{\cdot}{\lvl-1} = \parent(\anc{\cdot}{\lvl})$.
    In a similar fashion, we can write
        \begin{align*}
        \adist\brb{A_{t,2},\anc{\varfunc_{t}}{\lvl_t-1}(x_t)} 
        &= \adist\brb{\func_{t,2}(x_t),\anc{\func_{t,2}}{\lvl_t-1}(x_t)} \\
        &= \sum_{\lvl = \lvl_t}^{\nlvls} \adist\brb{\anc{\func_{t,2}}{\lvl}(x_t),\anc{\func_{t,2}}{\lvl-1}(x_t)} \\
        &\leq  \sum_{\lvl = \lvl_t}^{\nlvls} \prtcst 2^{-\lvl+1} \leq \prtcst 2^{-\lvl_t + 2} \,.
    \end{align*}
    Then, we can conclude by using the triangle inequality one more time, combined with the definition of $\adist$:
    \begin{align*}
        \abs{ \loss_t(A_{t,1}) - \loss_t(A_{t,2}) } &\leq \adist(A_{t,1},A_{t,2}) \\
        &\leq \adist\brb{A_{t,1},\anc{\varfunc_{t}}{\lvl_t-1}(x_t)} + \adist\brb{A_{t,2},\anc{\varfunc_{t}}{\lvl_t-1}(x_t)} \\
        &\leq \prtcst 2^{-\lvl_t + 3} \,.
    \end{align*}
\end{proof}

\begin{lemma}\label{lem:loss-diff-plus-b-range}
    Choosing $b_\lvl = \prtcst 2^{-\lvl+3}$ at every level guarantees that 
    \[
        0\leq \loss_t(A_{t,1}) - \loss_t(A_{t,2}) + b_{\lvl_t} \leq \prtcst 2^{-\lvl_t+4} \,.
    \]
\end{lemma}
\begin{proof}
    Since \Cref{lem:loss-diff-bound} gives that $\abs{ \loss_t(A_{t,1}) - \loss_t(A_{t,2}) } \leq \prtcst 2^{-\lvl_t + 3}$, the stated choice of $b_\lvl$ yields the lemma's statement.
\end{proof}

\begin{lemma}\label{lem:estnodeloss}
    For $\lvl \in [\nlvls]$ and $\func \in \expertree_\lvl$, define 
    \[
        \estnodeloss_{t,\lvl}( \func ) \coloneqq \frac{\indicator{ \bee_{t,1} = \func(x_t) \,,\, \bee_{t,2}  = \parent(\func)(x_t)\,,\, \lvl_t = \lvl } \brb{\loss_t(A_{t,1}) - \loss_t(A_{t,2}) + b_{\lvl_t}}} {\pr_t[ \bee_{t,1} = \func(x_t) \,,\, \bee_{t,2}  = \parent(\func)(x_t) \,,\, \lvl_t = \lvl]} \,.
    \]
    Then, for any $\func \in \expertree_{\nlvls}$, 
    \[
        \hat{\expertloss}_t(\func) =  \sum_{\lvl = 1}^{\nlvls} \estnodeloss_{t,\lvl}( \anc{\func}{\lvl} ) \,.
    \]
    
\end{lemma}
\begin{proof}
    For $\func \in \expertree_\nlvls$ and $\lvl \in [\nlvls]$, define  \[\event_t(\func,\lvl) \coloneqq \{ \bee_{t,1} = \anc{\func}{\lvl}(x_t) \,,\, \bee_{t,2}  = \anc{\func}{\lvl - 1}(x_t)   \}\,.\]
    Observe then that $\hat{\expertloss}_t(\func)$, as defined in \Cref{alg:2-point-feedback-experts}, can be written as
    \begin{align*}
         \hat{\expertloss}_t(\func) &= \frac{\indicator{ \event_t(\func,\lvl_t) } \brb{\loss_t(A_{t,1}) - \loss_t(A_{t,2}) + b_{\lvl_t}}} {\delta_{\lvl_t} \pr_t[ \event_t(\func,\lvl_t) \mid \lvl_t ]} \\
         &\hspace{0em}= \sum_{\lvl = 1}^{\nlvls} \frac{\indicator{ \event_t(\func,\lvl), \lvl_t = \lvl } \brb{\loss_t(A_{t,1}) - \loss_t(A_{t,2}) + b_{\lvl_t}}} {\delta_{\lvl} \pr_t[ \event_t(\func,\lvl) \mid \lvl_t = \lvl] } \\
         &\hspace{0em}= \sum_{\lvl = 1}^{\nlvls} \frac{\indicator{ \event_t(\func,\lvl), \lvl_t = \lvl } \brb{\loss_t(A_{t,1}) - \loss_t(A_{t,2}) + b_{\lvl_t}}} {\pr_t[\lvl_t = \lvl] \pr_t[ \event_t(\func,\lvl) \mid \lvl_t = \lvl]} \\
         &\hspace{0em}= \sum_{\lvl = 1}^{\nlvls} \frac{\indicator{ \event_t(\func,\lvl), \lvl_t = \lvl } \brb{\loss_t(A_{t,1}) - \loss_t(A_{t,2}) + b_{\lvl_t}}} {\pr_t[ \event_t(\func,\lvl), \lvl_t = \lvl]} \,,
    \end{align*} 
    where we have used that $\delta_\lvl = \pr_t(\lvl_t = \lvl)$.
    The lemma then follows from the definition of $\estnodeloss_{t,\lvl}(\cdot)$ and the fact that $\anc{\func}{\lvl-1} = \parent(\anc{\func}{\lvl})$.
\end{proof}

\begin{lemma}\label{lem:expectation-of-estnodeloss}
    For $\lvl \in [\nlvls]$ and $\func \in \expertree_\lvl$, it holds that 
    \[\E_t\bsb{ \estnodeloss_{t,\lvl}(\func) } = \E_t\bsb{\loss_t(A_{t,1}) - \loss_t(A_{t,2}) + b_{\lvl_t} \mid \bee_{t,1} = \func(x_t) \,,\, \bee_{t,2}  = \prtat{\func}{\lvl - 1}(x_t) \,,\, \lvl_t = \lvl} \,.\]
\end{lemma}
\begin{proof}
    Observe from the proof of \Cref{lem:estnodeloss} that we can write
    \begin{align*}
        \estnodeloss_{t,\lvl}( \func ) = \frac{\indicator{ \event_t(\func,\lvl), \lvl_t = \lvl } \brb{\loss_t(A_{t,1}) - \loss_t(A_{t,2}) + b_{\lvl_t}}} {\pr_t[ \event_t(\func,\lvl), \lvl_t = \lvl]} \,,
    \end{align*}
    where 
    \[\event_t(\func,\lvl) = \{ \bee_{t,1} = \anc{\func}{\lvl}(x_t) \,,\, \bee_{t,2}  = \anc{\func}{\lvl - 1}(x_t)   \} = \{ \bee_{t,1} = \func(x_t) \,,\, \bee_{t,2}  = \prtat{\func}{\lvl - 1}(x_t)   \} \,.\]
    The second equality above holds since $\func$ here is a member of $\expertree_\lvl$.
    Taking the expectation of the numerator yields 
    \begin{align*}
        &\E_t\lsb{\indicator{ \event_t(\func,\lvl), \lvl_t = \lvl } \brb{\loss_t(A_{t,1}) - \loss_t(A_{t,2}) + b_{\lvl_t}}}\\ 
        &\hspace{1em}= \pr_t\bsb{\event_t(\func,\lvl), \lvl_t = \lvl} \E_t\lsb{\indicator{ \event_t(\func,\lvl), \lvl_t = \lvl } \brb{\loss_t(A_{t,1}) - \loss_t(A_{t,2}) + b_{\lvl_t}} \mid \event_t(\func,\lvl), \lvl_t = \lvl} \\
        &\hspace{2.5em}+ \pr_t\Bsb{\overbar{\event_t(\func,\lvl)} \lor \lvl_t \neq \lvl} \E_t\lsb{\indicator{ \event_t(\func,\lvl), \lvl_t = \lvl } \brb{\loss_t(A_{t,1}) - \loss_t(A_{t,2}) + b_{\lvl_t}} \mid \overbar{\event_t(\func,\lvl)} \lor \lvl_t \neq \lvl} \\
        &\hspace{1em}= \pr_t\bsb{\event_t(\func,\lvl), \lvl_t = \lvl} \E_t\bsb{\loss_t(A_{t,1}) - \loss_t(A_{t,2}) + b_{\lvl_t} \mid \event_t(\func,\lvl), \lvl_t = \lvl} \,.
    \end{align*}
    The lemma then follows since the denominator of $\estnodeloss_{t,\lvl}( \func )$ is $\cF_{t-1}$-measurable.
\end{proof}

Recall that we have defined $\expertloss_t(\func) \coloneqq \loss_t(\func(x_t))$ for $\func \in \expertree_\nlvls$.
\begin{lemma}\label{lem:telescoping-nodeloss}
    For $\lvl \in [\nlvls]$ and $\func \in \expertree_\lvl$, define
    \begin{align*}
        \avgnodeloss_{t,\lvl}(\func) \coloneqq \E_t \bsb{ \loss_t(A_{t,1}) - \loss_t(A_{t,2}) + b_{\lvl_t} \mid \bee_{t,1} = \func(x_t)\,,\, \bee_{t,2} = \prtat{\func}{\lvl-1}(x_t)\,,\,  \lvl_t = \lvl  } \,.
    \end{align*}
    Then, any $\func, \func' \in \expertree_\nlvls$ satisfy that
    \begin{align*}%
        \expertloss_t(\func) - \expertloss_t(\func')
        = \sum_{\lvl \in [\nlvls]} \avgnodeloss_{t,\lvl}(\anc{\func}{\lvl}) - \sum_{\lvl \in [\nlvls]} \avgnodeloss_{t,\lvl}(\anc{\func'}{\lvl})  \,.
    \end{align*}
\end{lemma}
\begin{proof}
        Observe that
    \begin{align*}
        &\E_t \bsb{ \loss_t(A_{t,1}) \mid \bee_{t,1} = \func(x_t), \bee_{t,2} = \prtat{\func}{\lvl-1}(x_t),  \lvl_t = \lvl  } \\
        &\hspace{6em}= \E_t \bsb{ \loss_t(\func_{t,1}(x_t)) \mid \anc{\varfunc_t}{\lvl_t}(x_t) = \func(x_t), \anc{\varfunc_t}{\lvl_t-1}(x_t) = \prtat{\func}{\lvl-1}(x_t),  \lvl_t = \lvl  } \\
        &\hspace{6em}= \E_t \bsb{ \loss_t(\func_{t,1}(x_t)) \mid \anc{\varfunc_t}{\lvl_t}(x_t) = \func(x_t), \lvl_t = \lvl  } \\
        &\hspace{6em}= \E_t \bsb{ \loss_t(\func_{t,1}(x_t)) \mid \anc{\func_{t,1}}{\lvl}(x_t) = \func(x_t), \lvl_t = \lvl  } \\
        &\hspace{6em}= \E_t \bsb{ \loss_t(\func_{t,1}(x_t)) \mid \anc{\func_{t,1}}{\lvl}(x_t) = \func(x_t) } \,,
    \end{align*}
    where the first step uses the definitions of $A_{t,1}$, $\bee_{t,1}$, and $\bee_{t,2}$; the second uses the fact that knowing $\lvl_t$ and $\anc{\varfunc_t}{\lvl_t}(x_t)$, $\func_{t,1}$ is independent of $\anc{\varfunc_t}{\lvl_t-1}$;
    the third uses that $\anc{\varfunc_t}{\lvl_t}(x_t) = \anc{\func_{t,1}}{\lvl_t}(x_t)$; 
    and the last equality uses that $\func_{t,1} \ind \lvl_t \mid \cF_{t-1}$ as asserts \Cref{lem:same-marginals-independent-from-level}.
    In a similar manner,
    \begin{align*}
        &\E_t \bsb{ \loss_t(A_{t,2}) \mid \bee_{t,1} = \func(x_t), \bee_{t,2} = \prtat{\func}{\lvl-1}(x_t),  \lvl_t = \lvl  } \\
        &\hspace{6em}= \E_t \bsb{ \loss_t(\func_{t,2}(x_t)) \mid \anc{\varfunc_t}{\lvl_t}(x_t) = \func(x_t), \anc{\varfunc_t}{\lvl_t-1}(x_t) = \prtat{\func}{\lvl-1}(x_t),  \lvl_t = \lvl  } \\
        &\hspace{6em}= \E_t \bsb{ \loss_t(\func_{t,2}(x_t)) \mid \anc{\varfunc_t}{\lvl_t-1}(x_t) = \prtat{\func}{\lvl-1}(x_t), \lvl_t = \lvl  } \\
        &\hspace{6em}= \E_t \bsb{ \loss_t(\func_{t,2}(x_t)) \mid \anc{\func_{t,2}}{\lvl-1}(x_t) = \prtat{\func}{\lvl-1}(x_t), \lvl_t = \lvl  } \\
        &\hspace{6em}= \E_t \bsb{ \loss_t(\func_{t,2}(x_t)) \mid \anc{\func_{t,2}}{\lvl-1}(x_t) = \prtat{\func}{\lvl-1}(x_t)} \,,
    \end{align*}
    where the first step uses the definitions of $A_{t,2}$, $\bee_{t,1}$, and $\bee_{t,2}$; the second uses the fact that knowing $\lvl_t$ and $\anc{\varfunc_t}{\lvl_t-1}(x_t)$, $\func_{t,2}$ is independent of $\anc{\varfunc_t}{\lvl_t}$;
    the third uses that $\anc{\varfunc_t}{\lvl_t-1}(x_t) = \anc{\func_{t,2}}{\lvl_t-1}(x_t)$; 
    and the last equality uses that $\func_{t,2} \ind \lvl_t \mid \cF_{t-1}$ via \Cref{lem:same-marginals-independent-from-level}.
    Hence, it holds for any $\func \in \expertree_\nlvls$ that
    \begin{align*}
        \sum_{\lvl \in [\nlvls]} \avgnodeloss_{t,\lvl}(\anc{\func}{\lvl}) &= \sum_{\lvl \in [\nlvls]} \E_t \bsb{ \loss_t(A_{t,1}) \mid \bee_{t,1} = \anc{\func}{\lvl}(x_t), \bee_{t,2} = \anc{\func}{\lvl-1}(x_t),  \lvl_t = \lvl  } \\
        &\hspace{4em} - \E_t \bsb{ \loss_t(A_{t,2}) \mid \bee_{t,1} = \anc{\func}{\lvl}(x_t), \bee_{t,2} = \anc{\func}{\lvl-1}(x_t),  \lvl_t = \lvl  } + b_{\lvl} \\
        &\hspace{0em} = \sum_{\lvl \in [\nlvls]} \E_t \bsb{ \loss_t(\func_{t,1}(x_t)) \mid \anc{\func_{t,1}}{\lvl}(x_t) = \anc{\func}{\lvl}(x_t) }  \\
        &\hspace{4em}- \E_t \bsb{ \loss_t(\func_{t,2}(x_t)) \mid \anc{\func_{t,2}}{\lvl-1}(x_t) = \anc{\func}{\lvl-1}(x_t)} + b_{\lvl} \\
        &\hspace{0em} = \E_t \bsb{ \loss_t(\func_{t,1}(x_t)) \mid \anc{\func_{t,1}}{\nlvls}(x_t) = \anc{\func}{\nlvls}(x_t) } \\
        &\hspace{4em}- \E_t \bsb{ \loss_t(\func_{t,2}(x_t)) \mid \anc{\func_{t,2}}{0}(x_t) = \anc{\func}{0}(x_t)} + \sum_{\lvl \in [\nlvls]} b_{\lvl} \\
        &\hspace{0em} = \loss_t(\func(x_t)) - \E_t \bsb{ \loss_t(A_{t,2})} + \sum_{\lvl \in [\nlvls]} b_{\lvl} = \expertloss_t(\func) - \E_t \bsb{ \loss_t(A_{t,2})} + \sum_{\lvl \in [\nlvls]} b_{\lvl} \,,
    \end{align*}
    where the first equality holds since $\prtat{\anc{\func}{\lvl}}{\lvl-1} = \anc{\func}{\lvl-1}$;
    the second uses the previous two derivations;
    the third uses the telescoping property of the sum, which follows since $\pr_t(\func_{t,1}) = \pr_t(\func_{t,2})$ via \Cref{lem:same-marginals-independent-from-level};
    and the fourth uses that $\anc{\func}{\nlvls} = \func$ and $\anc{\func}{0} = \func_0$ for all $\func \in \expertree_\nlvls$.
    As a consequence, any $\func, \func' \in \expertree_\nlvls$ satisfy, as required, that
    \begin{align*}%
        \sum_{\lvl \in [\nlvls]} \avgnodeloss_{t,\lvl}(\anc{\func}{\lvl}) - \sum_{\lvl \in [\nlvls]} \avgnodeloss_{t,\lvl}(\anc{\func'}{\lvl}) = \expertloss_t(\func) - \expertloss_t(\func') \,.
    \end{align*}
\end{proof}

\begin{lemma}\label{lem:expectation-of-expertloss}
    For any $\func,\func' \in \expertree_\nlvls$, it holds that
    \begin{align*}
        \expertloss_t(\func) - \expertloss_t(\func') = \E_t\bsb{\hat{\expertloss}_t(\func) - \hat{\expertloss}_t(\func')} \,.
    \end{align*}
\end{lemma}
\begin{proof}
    Combining \Cref{lem:estnodeloss,lem:expectation-of-estnodeloss,lem:telescoping-nodeloss}, we get
    \begin{align*}
        \E_t[\hat{\expertloss}_t(\func)]- \E_t[\hat{\expertloss}_t(\func')] &=  \sum_{\lvl = 1}^{\nlvls} \E_t\bsb{\estnodeloss_{t,\lvl}( \anc{\func}{\lvl} )} - \sum_{\lvl = 1}^{\nlvls} \E_t\bsb{\estnodeloss_{t,\lvl}( \anc{\func'}{\lvl} )} \\
        &= \sum_{\lvl = 1}^{\nlvls} \avgnodeloss_{t,\lvl}( \anc{\func}{\lvl} ) - \sum_{\lvl = 1}^{\nlvls} \avgnodeloss_{t,\lvl}( \anc{\func'}{\lvl} ) \\
        &= \expertloss_t(\func) - \expertloss_t(\func') \,.
    \end{align*}
\end{proof}

\begin{proposition} \label{prop:generic-thm}
    Under \Cref{assump:function-tree},
    \Cref{alg:2-point-feedback-experts} with $\psi(p) \coloneqq \sum_{\lvl \in [\nlvls]} 1/\eta_{\lvl} \sum_{\func \in \expertree_\lvl} p[\func] \ln p[\func]$ and $b_\lvl = \prtcst 2^{-\lvl+3}$ satisfies that
    \begin{equation*}
        \reg{T}{2} \leq \sum_{\lvl =1}^{\nlvls} \frac{1}{{\eta}_{\lvl}}  \ln |\expertree_\lvl|  + \prtcst^2  \sum_{\lvl =1}^{\nlvls} \eta_{\lvl} \frac{2^{-2\lvl+8}} {\delta_\lvl} \sum_{a \in \actions_{\lvl-1}} |\suc_\lvl(a)| \, T + \apprxcst 2^{-\nlvls} T \,.
    \end{equation*}
\end{proposition}
\begin{proof}
    Recall that $\expertloss_t(\func) \coloneqq \loss_t(\func(x_t))$ for $\func \in \expertree_\nlvls$.
    By \Cref{assump:function-tree}, for every $\expert \in \experts$, there exists some $\func \in \expertree_\nlvls$ such that $\distfunc(\expert, \func) \leq \apprxcst 2^{-\nlvls}$. 
    So, since $\expertloss_t(\func) - \expertloss_t(\expert) \leq \distfunc(\expert, \func)$,
    \begin{align*}
        \sum_{t=1}^T \expertloss_t(\expert) 
        &= \sum_{t=1}^T \expertloss_t(\func) - \sum_{t=1}^T \expertloss_t(\func) - \expertloss_t(\expert) \\
        &\geq \sum_{t=1}^T \expertloss_t(\func) - \sum_{t=1}^T \distfunc(\func,\expert) \geq \sum_{t=1}^T \expertloss_t(\func) - \apprxcst 2^{-\nlvls} T \,.
    \end{align*}
    Hence,
    \[
        \min_{\func \in \expertree_\nlvls} \sum_{t=1}^T \expertloss_t(\func) \leq \apprxcst 2^{-\nlvls} T + \inf_{\expert \in \experts} \sum_{t=1}^T \expertloss_t(\expert) \,,
    \]
    implying that
    \begin{align*}
        \reg{T}{2} 
        &\leq \E\left[\sum_{t=1}^T \frac{1}{2}( \loss_t(A_{t,1}) + \loss_t(A_{t,2}) )\right] - \min_{\func \in \expertree_\nlvls} \sum_{t=1}^T \expertloss_t(\func)  + \apprxcst 2^{-\nlvls} T \,.
    \end{align*}
    Since $A_{t,1} = \func_{t,1}(x_t)$ and $A_{t,2} = \func_{t,2}(x_t)$, \Cref{lem:same-marginals-independent-from-level} implies that $\pr_t(A_{t,1}) = \pr_t(A_{t,2})$.
    Hence, 
    \begin{align*}
        \E \lsb{\frac{1}{2}( \loss_t(A_{t,1}) + \loss_t(A_{t,2}) )} = \E \bsb{ \loss_t(A_{t,1}) } \,.
    \end{align*}
    Observe that $\cF_{t}$ is the $\sigma$-algebra generated by $(\varfunc_s, \func_{s,1}, \func_{s,2}, \lvl_s)_{s \leq t}$.
    Note then that $p_t$ is measurable with respect to $\cF_{t-1}$.
    And since \Cref{lem:same-marginals-independent-from-level} also asserts that $\pr_t(\func_{t,1}=\cdot) =  p_t(\cdot)$, we get that
    \[
        \E_t \bsb{ \loss_t(A_{t,1}) } = \E_t \bsb{ \loss_t(\func_{t,1}(x_t)) } = \sum_{\func \in \expertree_\nlvls} p_t(\func) \loss_t(\func(x_t)) = \sum_{\func \in \expertree_\nlvls} p_t(\func) \expertloss_t(\func) \,.
    \]
    Thus,
    \begin{align*} %
        \reg{T}{2} \leq \E \Bbsb{\sum_{t=1}^T \sum_{\func \in \expertree_\nlvls} p_t(\func) \expertloss_t(\func) } - \min_{\func' \in \expertree_\nlvls} \sum_{t=1}^T \expertloss_t(\func') + \apprxcst 2^{-\nlvls} T \,.
    \end{align*}
    So, if we fix $\func^* \in \argmin_{\func' \in \expertree_{\nlvls}}\ \sum_{t=1}^T \expertloss_t(\func')$ and apply \Cref{lem:expectation-of-expertloss} we get that
    \begin{align}
        \reg{T}{2} &\leq \E \Bbsb{\sum_{t=1}^T \sum_{\func \in \expertree_\nlvls} p_t(\func) \brb{\expertloss_t(\func) - \expertloss_t(\func^*)} }  + \apprxcst 2^{-\nlvls} T \nonumber\\
        \label{eq:reg-decomp-experts-estimated-losses}
        &= \E \Bbsb{\sum_{t=1}^T \sum_{\func \in \expertree_\nlvls} p_t(\func) \brb{\hat{\expertloss}_t(\func) - \hat{\expertloss}_t(\func^*)} }  + \apprxcst 2^{-\nlvls} T \,.
    \end{align}
    Note that the second step used that $p_t$ is $\cF_{t-1}$-measurable.    
    Recall that in \Cref{lem:estnodeloss}, we have defined
    \[
        \estnodeloss_{t,\lvl}( \func ) \coloneqq \frac{\indicator{ \bee_{t,1} = \func(x_t) \,,\, \bee_{t,2}  = \parent(\func)(x_t)\,,\, \lvl_t = \lvl } \brb{\loss_t(A_{t,1}) - \loss_t(A_{t,2}) + b_{\lvl_t}}} {\pr_t[ \bee_{t,1} = \func(x_t) \,,\, \bee_{t,2}  = \parent(\func)(x_t) \,,\, \lvl_t = \lvl]} \,.
    \]
    for $\lvl \in [\nlvls]$ and $\func \in \expertree_\lvl$.
    Now, since
    \begin{itemize}[nosep]
        \item $\displaystyle p_{t+1} = \argmin_{p \in \cP_{\expertree,\nlvls}} \ban{\summ_{s \leq t} \hat{\expertloss}_s,p} + \psi(p)$ for all rounds $t$,
        \item $\displaystyle \hat{\expertloss}_t(\func) =  \sum_{\lvl = 1}^{\nlvls} \estnodeloss_{t,\lvl}( \anc{\func}{\lvl} )$ for all $\func \in \expertree_{\nlvls}$ (\Cref{lem:estnodeloss}), and
        \item $\displaystyle \estnodeloss_{t,\lvl}( \func ) \geq 0$ for all $\lvl \in [\nlvls]$ and $\func \in \expertree_{\lvl}$ (\Cref{lem:loss-diff-plus-b-range} combined with the definition of $\estnodeloss_{t,\lvl}$);
    \end{itemize}
    we can use \Cref{lem:ftrl-nested-shannon-generic} to assert that
        \begin{align*}
            &\sum_{t=1}^T \sum_{\func \in \expertree_\nlvls} p_t(\func) \brb{\hat{\expertloss}_t(\func) - \hat{\expertloss}_t(\func^*)}  \\
            &\hspace{5em}\leq \sum_{\lvl =1}^{\nlvls} \frac{1}{{\eta}_{\lvl}}  \sum_{\func \in \expertree_{\lvl}} p_{1}[\func] \ln(1/p_{1}[\func])  + \sum_{t=1}^T  \sum_{\lvl =1}^{\nlvls} \eta_{\lvl} \sum_{\func \in \expertree_{\lvl}} p_{t}[\func] \brb{\estnodeloss_{t,\lvl}(\func)}^2 \,.
        \end{align*}
    Notice that for any $\lvl \in [\nlvls]$,
    \begin{align*}
        &\E_t \lsb{ \sum_{\func \in \expertree_{\lvl}} p_{t}[\func] \brb{\estnodeloss_{t,\lvl}(\func)}^2 } \\
        &\hspace{3em}= \sum_{\func \in \expertree_{\lvl}} p_{t}[\func] \E_t \lsb{\frac{\indicator{ \bee_{t,1} = \func(x_t) \,,\, \bee_{t,2}  = \parent(\func)(x_t)\,,\, \lvl_t = \lvl } \brb{\loss_t(A_{t,1}) - \loss_t(A_{t,2}) + b_{\lvl_t}}^2} {\pr_t[ \bee_{t,1} = \func(x_t) \,,\, \bee_{t,2}  = \parent(\func)(x_t) \,,\, \lvl_t = \lvl]^2} } \\
        &\hspace{3em}\leq \prtcst^2 \sum_{\func \in \expertree_{\lvl}} p_{t}[\func] \E_t \lsb{\frac{\indicator{ \bee_{t,1} = \func(x_t) \,,\, \bee_{t,2}  = \parent(\func)(x_t)\,,\, \lvl_t = \lvl }  2^{-2\lvl_t+8}} {\pr_t[ \bee_{t,1} = \func(x_t) \,,\, \bee_{t,2}  = \parent(\func)(x_t) \,,\, \lvl_t = \lvl]^2} } \\
        &\hspace{3em}= \prtcst^2 \sum_{\func \in \expertree_{\lvl}} p_{t}[\func] \E_t \lsb{\frac{\indicator{ \bee_{t,1} = \func(x_t) \,,\, \bee_{t,2}  = \parent(\func)(x_t)\,,\, \lvl_t = \lvl }  2^{-2\lvl+8}} {\pr_t[ \bee_{t,1} = \func(x_t) \,,\, \bee_{t,2}  = \parent(\func)(x_t) \,,\, \lvl_t = \lvl]^2} } \\
        &\hspace{3em}= \prtcst^2 2^{-2\lvl+8} \sum_{\func \in \expertree_{\lvl}}  \frac{p_{t}[\func] } {\pr_t[ \bee_{t,1} = \func(x_t) \,,\, \bee_{t,2}  = \parent(\func)(x_t) \,,\, \lvl_t = \lvl]} \,,
    \end{align*}
    where we have used in the first equality the definition of $\estnodeloss_{t,\lvl}$ and the fact that $p_t$ is $\cF_{t-1}$-measurable; and in the inequality the bound of \Cref{lem:loss-diff-plus-b-range}.
    Going further,
    \begin{align*}
        &\sum_{\func \in \expertree_{\lvl}}  \frac{p_{t}[\func] } {\pr_t[ \bee_{t,1} = \func(x_t) \,,\, \bee_{t,2}  = \parent(\func)(x_t) \,,\, \lvl_t = \lvl]}  \\
        &\hspace{3em}=\sum_{\func \in \expertree_{\lvl}}  \frac{p_{t}[\func] } {\pr_t[ \anc{\varfunc_{t}}{\lvl_t}(x_t) = \func(x_t) \,,\, \anc{\varfunc_{t}}{\lvl_t-1}(x_t)  = \parent(\func)(x_t) \,,\, \lvl_t = \lvl]}  \\
        &\hspace{3em}=\sum_{\func \in \expertree_{\lvl}}  \frac{p_{t}[\func] } {\pr_t[ \anc{\varfunc_{t}}{\lvl}(x_t) = \func(x_t) \,,\, \anc{\varfunc_{t}}{\lvl-1}(x_t)  = \parent(\func)(x_t) \,,\, \lvl_t = \lvl]}  \\
        &\hspace{3em}=\sum_{\func \in \expertree_{\lvl}}  \frac{p_{t}[\func] } {\pr_t[ \anc{\varfunc_{t}}{\lvl}(x_t) = \func(x_t) \,,\, \anc{\varfunc_{t}}{\lvl-1}(x_t)  = \parent(\func)(x_t)] \pr_t[\lvl_t = \lvl]}\\
        &\hspace{3em}= \delta^{-1}_\lvl \sum_{\func \in \expertree_{\lvl}}  \frac{p_{t}[\func] } {\pr_t[ \anc{\varfunc_{t}}{\lvl}(x_t) = \func(x_t) \,,\, \anc{\varfunc_{t}}{\lvl-1}(x_t)  = \parent(\func)(x_t)]}\\
        &\hspace{3em}= \delta^{-1}_\lvl \sum_{a \in \actions_{\lvl-1}} \sum_{b \in \suc_{\lvl}(a)}  \frac{\sum_{\func \in \expertree_{\lvl}} \indicator{\func(x_t) = b, \prtat{\func}{\lvl-1}(x_t) = a} p_{t}[\func] } {\pr_t[ \anc{\varfunc_{t}}{\lvl}(x_t) = b \,,\, \anc{\varfunc_{t}}{\lvl-1}(x_t)  = a] } \leq \delta^{-1}_\lvl \sum_{a \in \actions_{\lvl-1}} |\suc_\lvl(a)| \,,
    \end{align*}
    where the first equality follows from the definitions of $\bee_{t,1}$ and $\bee_{t,2}$; the third uses that $\varfunc_t$ and $\lvl_t$ are sampled independently; the fourth uses that $\pr_t[\lvl_t = \lvl] = \delta_\lvl$; the fifth uses that $\prtat{\func}{\lvl-1}(x_t) \in \actions_{\lvl-1}$ and $\func(x_t) \in \suc_{\lvl}(\prtat{\func}{\lvl-1}(x_t))$; and the inequality uses that $\pr_t(\varfunc_t = \cdot) = p_t(\cdot)$, which implies that
    \begin{align*}
        &\pr_t[ \anc{\varfunc_{t}}{\lvl}(x_t) = b , \anc{\varfunc_{t}}{\lvl-1}(x_t)  = a] \\
        &\hspace{6em}= \sum_{\func' \in \expertree_\nlvls} p_t(f') \indicator{ \anc{\func'}{\lvl}(x_t) = b , \anc{\func'}{\lvl-1}(x_t)  = a } \\
        &\hspace{6em}= \sum_{\func \in \expertree_\lvl} \indicator{ \func(x_t) = b , \parent(\func)(x_t)  = a } \sum_{\func' \in \expertree_\nlvls} \indicator{\func = \anc{\func'}{\lvl}} p_t(f') \\
        &\hspace{6em}= \sum_{\func \in \expertree_\lvl} \indicator{ \func(x_t) = b , \parent(\func)(x_t)  = a } p_t[\func] \,.
    \end{align*}
    Combining the above with the fact that $\sum_{\func \in \expertree_{\lvl}} p_{t}[\func] \ln(1/p_{t}[\func]) \leq \ln |\expertree_\lvl|$ for all $\lvl \in [\nlvls]$ yields that
    \begin{align*}
        & \E \lsb{\sum_{t=1}^T \sum_{\func \in \expertree_\nlvls} p_t(\func) \brb{\hat{\expertloss}_t(\func) - \hat{\expertloss}_t(\func^*)} } \leq\sum_{\lvl =1}^{\nlvls} \frac{1}{{\eta}_{\lvl}}  \ln |\expertree_\lvl|  + \prtcst^2  \sum_{\lvl =1}^{\nlvls} \eta_{\lvl} \frac{2^{-2\lvl+8}} {\delta_\lvl} \sum_{a \in \actions_{\lvl-1}} |\suc_\lvl(a)| \, T \,,
    \end{align*}
    which concludes the proof after combining this with \eqref{eq:reg-decomp-experts-estimated-losses}.
\end{proof}

\subsection*{Auxiliary results}
The following Lemma is a direct adaptation of Lemma 16 in \cite{eldowa26a} to the nested negative entropy regularizer.
\begin{lemma}
\label{lem:ftrl-nested-shannon-generic}
    Let $(\eta_{t,\lvl})_{t \geq 1, \lvl \in [\nlvls]}$ be a sequence of positive numbers.
    For every $t \geq 1$, let $\psi_t$ be given by
    $
        \psi_t(p) = \sum_{\lvl \in [\nlvls]} \frac{1}{\eta_{t,\lvl}} \sum_{\func \in \expertree_\lvl} p[\func] \ln p[\func]
    $
    for $p \in \cP_{\expertree,\nlvls}$.
    Let $(\lossvec_t)_t$ be a sequence of loss functions mapping $\expertree_\nlvls$ to $\R$, and define \[
        p_1 \coloneqq \argmin_{p \in \cP_{\expertree,\nlvls}} \psi_1(p) \quad \text{and for}\:\: t \geq 2, \:\:
        p_t \coloneqq \argmin_{p \in \cP_{\expertree,\nlvls}} \ban{\summ_{s \leq t-1} \lossvec_s,p} + \psi_t(p) \,.
    \]
    For each $t \geq 1$ and $\lvl \in [\nlvls]$, fix a function $\lossvec_{t,\lvl} \colon \expertree_\lvl \rightarrow \R$ such that
    \[
        \lossvec_t(\func) =  \sum_{\lvl\in[\nlvls]} \lossvec_{t,\lvl}(\anc{\func}{\lvl}) \:\:\forall \func \in \expertree_{\nlvls} \quad \text{and} \quad \lossvec_{t,\lvl}(\func) \geq - \frac{1}{\eta_{t,\lvl}} \:\:\forall \lvl  \in [\nlvls] \,,\, \func \in \expertree_\lvl \,.
    \]
    Then, it holds for horizon $T \geq 1$ and comparator $q \in \cP_{\expertree,\nlvls}$ that
    \begin{align*}
        \sum_{t=1}^T \inprod{\lossvec_t, p_t - q} \leq \sum_{t=1}^{T+1} \sum_{\lvl \in [\nlvls]} \tilde{\eta}_{t,\lvl}  \sum_{\func \in \expertree_\lvl} p_t[\func] \ln(1/p_t[\func])  
        + \sum_{t=1}^T  \sum_{\lvl \in [\nlvls]} \eta_{t,\lvl} \sum_{\func \in \expertree_\lvl} p_t[\func] \lossvec_{t,\lvl}(\func) ^2 \,,
    \end{align*}
    where $\tilde{\eta}_{1,\lvl} = \eta_{1,\lvl}^{-1}$ and for $t\geq 2$, $\tilde{\eta}_{t,\lvl} = \eta_{t,\lvl}^{-1} - \eta_{t-1,\lvl}^{-1} \,.$
\end{lemma}

\section{Proofs of Section \ref{sec:concrete_rate}}\label{app:concrete}
The following is a proof of \Cref{thm:concrete}, which we restate below.

\thmconcrete*
\begin{proof}
    Seeing as $\expertree_\nlvls$ is a $\extracst 2^{-\nlvls+1}$-cover of $\experts$ and $\distfunc(\func, \parent(\func)) \leq 3 \extracst 2^{-\lvl}$ for $\lvl < \nlvls$ and $\func \in \expertree_{\lvl+1}$; \Cref{assump:function-tree} is then satisfied with $\apprxcst = 2 \extracst$ and $\prtcst = 3 \extracst$.
    Using the tuning of \Cref{thm:generic-thm} hence yields that
    \begin{equation*}
        \reg{T}{2} \leq 96 \extracst \lrb{ \sum_{\lvl =1}^{\nlvls} 2^{-2 \lvl / 3} \Bbrb{\ln \brb{|\expertree_\lvl|} \sum_{a \in \actions_{\lvl-1}} |\suc_\lvl(a)|}^{1/3} }^{3/2} \sqrt{T} +  \extracst 2^{-\nlvls+1} T \,.
    \end{equation*}
    Now, fix $\lvl \in [\nlvls]$ and $a \in \actions_{\lvl-1}$, and recall that 
    \begin{align*} 
        \suc_{\lvl}(a) \coloneqq \Big\{ b \in \actions_{\lvl} \mid \exists x \in \contexts ,\, \func \in \expertree_{\lvl}\:\:\mathrm{s.t.}\:\:
        a =  \parent(\func)(x) \:\: \text{and}  \:\: b = \func(x) \Big\} \,.
    \end{align*}
    Since any $\func \in \expertree_{\lvl}$ satisfies $\distfunc(\func,\parent(\func)) \leq 6 \extracst 2^{-\lvl}$, it holds for any $b \in \suc_{\lvl}(a)$ that $\adist(a,b) \leq 6 \extracst 2^{-\lvl}$.
    So, $\suc_\lvl(a) \subseteq \ball(a; 6 \extracst 2^{-\lvl})$.
    Also, since $\suc_\lvl(a) \subseteq \actions_\lvl$ and $\actions_\lvl$ is a $\extracst 2^{-\lvl}$-packing, it must be that all distinct $b,b' \in \suc_\lvl(a)$ satisfy $\adist(b,b') > \extracst 2^{-\lvl}$.
    This implies that any ball of radius $\extracst 2^{-\lvl-1}$ can only contain one member of $\suc_\lvl(a)$.
    The upshot of this is that
    \[
        |\suc_\lvl(a)| \leq \covering\brb{ \ball(a; 6 \extracst 2^{-\lvl}) ;  \extracst 2^{-\lvl-1}} \,.
    \]
    Hence,
    \begin{align*}
        \covering\brb{ \ball(a; 6 \extracst 2^{-\lvl}) ;  \extracst 2^{-\lvl-1}} &\leq \covering\brb{ \ball(a; 6 \extracst 2^{-\lvl}) ;  6 \extracst 2^{-\lvl-4}} \\
        &\leq \prod_{i=0}^{3} \sup_{c \in \actions} \covering\brb{ \ball(c; 6 \extracst 2^{-\lvl-i}) ;  6 \extracst 2^{-\lvl-i-1}} \leq 2^{4 \dbldim} \,,
    \end{align*}
    where the last inequality uses the definition of $\dbldim$.
    Since this bound depends on neither $a$ nor $\lvl$, we get that
    \[
        \max_{\lvl \in [\nlvls], a \in \actions_{\lvl-1}} |\suc_\lvl(a)| \leq 2^{4 \dbldim} \,.
    \]
    Combining this with the fact that $\ln |\expertree_\lvl| \leq  \ln \babs{\covering(\experts; \extracst 2^{-\lvl})} = \metentropy(\experts; \extracst 2^{-\lvl}) \leq \funcst 2^{\lvl \fundim}$ and that $|\actions_{\lvl-1}| \leq \covering(\actions; \extracst 2^{-\lvl}) \leq \covcst 2^{\lvl\covdim} $; we arrive at
    \begin{equation*}
        \reg{T}{2} \leq 96 \extracst \sqrt{\covcst \funcst} \, 4^{\dbldim} \lrb{ \sum_{\lvl =1}^{\nlvls} 2^{(\covdim + \fundim-2) \lvl / 3} }^{3/2} \sqrt{T} +  \extracst 2^{-\nlvls+1} T \,.
    \end{equation*}
    For brevity, let $\alpha \coloneqq \frac{\covdim + \fundim-2}{3}$. 
    Notice that
    \begin{align*}
        \sum_{\lvl=1}^\nlvls 2^{\alpha \lvl} 
        = \frac{1 - 2^{\alpha (\nlvls + 1)}}{1 - 2^{\alpha}} - 1 
        = 2^\alpha \frac{1 - 2^{\alpha \nlvls}}{1 - 2^{\alpha}} &\leq \begin{cases}
            \lrb{1-2^{-\alpha}}^{-1} 2^{\alpha \nlvls} &\text{if } \alpha > 0\\
            \lrb{2^{-\alpha}-1}^{-1} &\text{if } \alpha < 0\\
            \nlvls &\text{if } \alpha=0
        \end{cases} \\
        &\leq \min\bcb{ \abs{1-2^{-\alpha}}^{-1} , \nlvls } \max\{1, 2^{\alpha \nlvls}\} \,.
    \end{align*}
    Now, if we set
    $\nlvls = \bfl{\frac{\log_2(T)}{(\covdim + \fundim) \lor 2 }}$, then $2^{-\nlvls+1} T \leq 4 T^{1-\frac{1}{(\covdim + \fundim) \lor 2 }}$ and
    \[
        \lrb{ \sum_{\lvl =1}^{\nlvls} 2^{(\covdim + \fundim-2) \lvl / 3} }^{3/2} \sqrt{T} \leq \bbrb{ \babs{1-2^{-\frac{\covdim + \fundim-2}{3}}}^{-3/2} \land \bbrb{ \frac{\log_2(T)}{(\covdim + \fundim) \lor 2 }}^{3/2} } T^{1-\frac{1}{(\covdim + \fundim) \lor 2 }} \,,
    \]
    which implies the sought result.
\end{proof}

\section{Efficient Implementation: More Details}\label{app:efficient}

\subsection{Structural properties}\label{app:efficient:tree-props}
We show here that the tree $\cT(\expertree_{0:\nlvls})$ constructed in \Cref{sec:efficient} satisfies \Cref{assump:function-tree}.
\begin{lemma} \label{lem:efficient:parent-child}
    For $\config \in \expertree_\lvl$ and $\config' \in \children(\config)$, $\norm{\func_{\config}(x) - \func_{\config'}(x)}_\infty \leq 3 \cdot 2^{-\lvl-2} $ for all $x \in \contexts$.
\end{lemma}
\begin{proof}
    Fix $x \in \contexts$ and let $\vertices = \seq(x)$.
    Then,
    \begin{align*}
        \norm{\func_{\config}(x) - \func_{\config'}(x)}_\infty &= \bno{\prof_{\vertices_{\upto \lvl}} - \prof'_{\vertices_{\upto \lvl+1}}}_\infty = \bno{\prof'_{\vertices_{\upto \lvl}} - \prof'_{\vertices_{\upto \lvl+1}}}_\infty
        \leq 3 \cdot 2^{-\lvl-2} \,,
    \end{align*}
    where the first equality used that $\config' \in \children(\config)$ and the second used that $\vertices_{\upto \lvl} = \parent(\vertices_{\upto \lvl+1})$.
\end{proof}

\begin{lemma} \label{lem:efficient:dense-levels}
    For every $\lvl \geq 0$ and $\expert \in \experts$, there exists some $\config \in \expertree_\lvl$ such that at every $\cnode \in \cnodes_\lvl$, $\bno{\expert(\ctx(\vertices)) -\prof_\vertices }_\infty \leq 2^{-\lvl-1}$.
\end{lemma}
\begin{proof}
    Fix $\expert \in \experts$ throughout the proof.
    We demonstrate this by induction.
    At level $0$, this is trivially satisfied by the root $\config = (\prof_{\rootvertex})$ for every $\expert$ since $\prof_{\rootvertex} = 2^{-1} \ones_\adim$ and $\image(\expert) \subseteq [0,1]^\adim$.
    Next, fix $\lvl \in \{0,\dots,\nlvls-1\}$ and assume that there exists $\config \in \expertree_\lvl$ such that $\bno{\expert(\ctx(\vertices)) -\prof_\vertices }_\infty \leq 2^{-\lvl-1}$ at every $\cnode \in \cnodes_\lvl$.
    Now fix some $\vertices \in \vertexset_{\lvl+1}$.
    It holds that 
    \[\bno{\expert(\ctx(\parent(\vertices))) - \expert(\ctx(\vertices))}_\infty \leq \bno{\ctx(\parent(\vertices)) - \ctx(\vertices)}_\infty \leq 2^{-\lvl-2} \,.\]
    Hence,
    \begin{align*}
    &\bno{\expert(\ctx(\vertices)) - \prof_{\parent(\vertices)} }_\infty  = \bno{\expert(\ctx(\parent(\vertices))) - \prof_{\parent(\vertices)} + \expert(\ctx(\vertices)) - \expert(\ctx(\parent(\vertices))) }_\infty \leq 3 \cdot 2^{-\lvl-2} \,,
    \end{align*}
    implying that $\expert(\ctx(\vertices)) -\prof_{\parent(\vertices)} \in [-3\cdot 2^{-\lvl-2},3\cdot 2^{-\lvl-2}]^\adim$.
    By the definition of $\children(\config)$,
    \[
        \{(\prof'_\vertices)_{\vertices \in \vertexset_{\lvl+1}} \mid \config' \in \children(\config)\} = \bigtimes_{\vertices \in \vertexset_{\lvl+1}}  \brb{\prof_{\parent(\vertices)}  + 2^{-\lvl-2} \cdot \{-3,-1,1,3\}^\adim }\,.
    \]
    So, we can choose $\config' \in \children(\config)$ such that for every $\vertices \in \vertexset_{\lvl+1}$,
    \begin{align*}
        \bno{\expert(\ctx(\vertices)) -\prof'_{\vertices} }_\infty = \bno{\expert(\ctx(\vertices)) - \prof_{\parent(\vertices)} + \prof_{\parent(\vertices)} - \prof'_{\vertices} }_\infty \leq 2^{-\lvl-2} \,,
    \end{align*}
    which concludes the proof.
\end{proof}

For $\lvl \geq 0$ and $x \in \contexts$, we will use $\rep{x}{\lvl}$ as a shorthand for $\ctx(\seq_{\upto\lvl}(x))$, which is the representative context of $x$ at level $\lvl$.
\begin{lemma} \label{lem:efficient:covering-levels}
    For every $\lvl \geq 0$ and $\expert \in \experts$, there exists $\config \in \expertree_\lvl$ such that $\norm{\expert(x) - \func_\config(x)}_\infty \leq 2^{-\lvl}$ for all $x \in \contexts$.
\end{lemma}
\begin{proof}
    By \Cref{lem:efficient:dense-levels}, we can find some $\config \in \expertree_\lvl$ such that $\bno{\expert(\rep{x}{\lvl}) - \prof_{\seq_{\upto \lvl}(x) } }_\infty \leq 2^{-\lvl-1}$ for all $x \in \contexts$.
    Thus, for any $x \in \contexts$,
    \begin{align*}
        \norm{\expert(x) - \func_\config(x)}_\infty &\leq \norm{\expert(x) - \expert(\rep{x}{\lvl})}_\infty + \bno{\expert(\rep{x}{\lvl}) - \prof_{\seq_{\upto \lvl}(x) }}_\infty \\
        &\leq  \norm{x - \rep{x}{\lvl}}_\infty + 2^{-\lvl-1} \leq 2^{-\lvl-1} + 2^{-\lvl-1} = 2^{-\lvl} \,.
    \end{align*}
\end{proof}

\begin{lemma} \label{lem:efficient:satisfying-assump}
    The tree $\cT(\expertree_{0:\nlvls})$ constructed in \Cref{sec:efficient} satisfies \Cref{assump:function-tree} with $\apprxcst = \losslip$ and $\prtcst = 3 \cdot 2^{-2} \losslip$.   
\end{lemma}
\begin{proof}
    This follows from \Cref{lem:efficient:covering-levels,lem:efficient:parent-child}, combined with the fact that $\adist(a,b) = \losslip \norm{a-b}_\infty$ for $a,b \in \actions$.
\end{proof}

\subsection{Decomposing the sampling distribution}\label{app:efficient:dec}
For completeness, the following lemma provides a standard result regarding the minimization, over the simplex, of a linear function minus the entropy function $H(p) \coloneqq \sum_{i=1}^d p(i) \ln (1/p(i))$.
\begin{lemma} \label{lem:efficient:exp-closed-form}
    Fix any $y \in \R^d$, and define
    $\displaystyle
        p^* = \argmin_{p \in \Delta_d} \inprod{p, y} - H(p) \,.
    $ 
    Then,
    \begin{align*}
        p^*(i) = \frac{e^{-y(i)}}{\sum_{j=1}^d e^{-y(j)}} \qquad \text{and} \qquad \inprod{p^*, y} - H(p^*) = - \ln \lrb{ \sum_{j=1}^d e^{-y(j)}} \,.
    \end{align*}
\end{lemma}

The next result provides a hierarchical version of this result when using the nested negative entropy regularizer over $\expertree_\nlvls$.
Recall that $\cP_{\expertree,\nlvls}$ denotes the space of probability distributions over $\expertree_\nlvls$.
Also, for $p \in \cP_{\expertree,\nlvls}$, $\lvl \in \{0,\dots,\nlvls\}$, and $\config \in \expertree_{\lvl}$, recall that 
\begin{align*}
    p[\config] \coloneqq \sum_{\config' \in \expertree_\nlvls} \indicator{ \config = \anc{\config'}{\lvl} } p(\config') \,.
\end{align*}
When this pertains to $p_t$ at some round $t$, we use the notation $p_{t,\lvl}$.
For $\lvl \in \{0,\dots,\nlvls-1\}$, $\config \in \expertree_\lvl$, and $p \in \cP_{\expertree,\nlvls}$ such that $p[\config] > 0$, let $\cond_{\config}^p$ be a conditional distribution over $\children(\config)$ given by
\[
    \cond_{\config}^{p}(\config') = \frac{p[\config']}{p[\config]} \quad \forall  \config' \in \children(\config) \,.
\]
Thus, for any $\config \in \expertree_{\nlvls}$ with $p(\config) > 0$, we can, since $p[\config_0] = 1$, write
\[
    p(\config) = \prod_{\lvl \in \{0,\dots,\nlvls-1\}} \cond_{\ancfun_{\lvl}(\config)}^{p}(\ancfun_{\lvl+1}(\config)) \,. 
\]

\begin{lemma} \label{lem:efficient:exp-levels-closed-form}
    Fix any $\genericloss \colon \expertree_\nlvls \rightarrow \R$
    and let $\displaystyle q = \argmin_{p \in \cP_{\expertree,\nlvls}} \lan{p,\genericloss} + \psi(p)$, where
    \[
    \psi(p) \coloneqq \sum_{\lvl \in [\nlvls]} \frac{1}{\eta_{\lvl}} \sum_{\config \in \expertree_\lvl} p[\config] \ln p[\config] \,.
    \]
    Then, for $\lvl \in \{0,\dots,\nlvls-1\}$ and $\config \in \expertree_\lvl$,
    \[  \cond^{q}_{\config}(\config') =  \frac{\exp\brb{- \etas^{-1}_{\lvl+1} \genericloss_{\lvl+1}(\config')}}{\sum_{\config'' \in \children(\config)}  \exp\brb{-\etas^{-1}_{\lvl+1} \genericloss_{\lvl+1}(\config'')}} = \frac{\exp\brb{- \etas^{-1}_{\lvl+1} \genericloss_{\lvl+1}(\config')}}{  \exp\brb{-\etas^{-1}_{\lvl+1} \genericloss_{\lvl}(\config)}} \quad \forall \config' \in \children(\config) \,,\]
    where $\etas_\lvl = \sum_{\lvl' \geq \lvl} {1}/{\eta_{\lvl'}}$, and $\genericloss_\lvl \colon \expertree_\lvl \rightarrow \R$ is given by
    \[
        \genericloss_\lvl(\config) = - \etas_{\lvl+1} \ln \lrb{ \sum_{\config' \in \children(\config)} \exp\brb{- \etas^{-1}_{\lvl+1} \genericloss_{\lvl+1}(\config')}} 
    \]
    for $\config \in \expertree_{\lvl}$, with $\genericloss_\nlvls = \genericloss$.
\end{lemma}
\begin{proof}
    Observe that a distribution $p \in \cP_{\expertree,\nlvls}$ with positive entries can be identified by the tuple $\brb{ (p[\config])_{\config \in \expertree_{\lvl-1}}, (\cond^p_{\config})_{\config \in \expertree_{\lvl-1}} }$; that is the distribution it induces over $\expertree_{\nlvls-1}$ and the conditional distribution of each member of $\expertree_{\nlvls-1}$ over its children in $\expertree_{\nlvls}$.
    Instead of optimizing over $p$ directly, one can optimize separately over these two independent components.
    To leverage this, we rewrite both terms of our objective with this goal in mind:
    \begin{align*}
        \lan{p,\genericloss} &= \sum_{\config \in \expertree_\nlvls} p(\config) \genericloss(\config) \\
        &= \sum_{\config' \in \expertree_{\nlvls-1}} \sum_{\config \in \children(\config')} p(\config) \genericloss(\config) \\
        &= \sum_{\config' \in \expertree_{\nlvls-1}} p[\config'] \sum_{\config \in \children(\config')} \frac{p(\config)}{p[\config']} \genericloss(\config) \\
        &= \sum_{\config' \in \expertree_{\nlvls-1}} p[\config']  \lan{\cond^p_{\config'},\genericloss} \,.
    \end{align*}
    Similarly,
    \begin{align*}
        \sum_{\config \in \expertree_\nlvls} p(\config) \ln p(\config) 
        &= \sum_{\config' \in \expertree_{\nlvls-1}} \sum_{\config \in \children(\config')} p(\config) \ln p(\config) \\
        &= \sum_{\config' \in \expertree_{\nlvls-1}} p[\config'] \sum_{\config \in \children(\config')} \frac{p(\config)}{p[\config']} \ln p(\config) \\
        &= \sum_{\config' \in \expertree_{\nlvls-1}} p[\config'] \sum_{\config \in \children(\config')} \frac{p(\config)}{p[\config']} \ln \frac{p(\config)}{p[\config']} \\
        &\hspace{10em}+ \sum_{\config' \in \expertree_{\nlvls-1}} p[\config'] \sum_{\config \in \children(\config')} \frac{p(\config)}{p[\config']} \ln p[\config'] \\
        &= \sum_{\config' \in \expertree_{\nlvls-1}} p[\config'] \sum_{\config \in \children(\config')} \cond^p_{\config'}(\config) \ln \cond^p_{\config'}(\config) + \sum_{\config' \in \expertree_{\nlvls-1}} p[\config'] \ln p[\config'] \,.
    \end{align*}
    In combination, these two derivations yield that
    \begin{align*}
        &\lan{p,\genericloss} + \psi(p) = \sum_{\lvl \in [\nlvls-2]} \frac{1}{\eta_{\lvl}} \sum_{\config \in \expertree_\lvl} p[\config] \ln p[\config] + \bbrb{\frac{1}{\eta_{\nlvls-1}} + \frac{1}{\eta_{\nlvls}}} \sum_{\config \in \expertree_{\nlvls-1}} p[\config] \ln p[\config] \\
        &\hspace{8em}+ \sum_{\config' \in \expertree_{\nlvls-1}} p[\config'] \Bbrb{  \lan{\cond^p_{\config'},\genericloss} +  \frac{1}{\eta_{\nlvls}} \sum_{\config \in \children(\config')} \cond^p_{\config'}(\config) \ln \cond^p_{\config'}(\config) } \,.
    \end{align*}
    Notice now that the quantity within the round brackets can be optimized in $\cond^p_{\config'}$ independently of the rest of the expression.
    Notice too that it takes the form of the program in \Cref{lem:efficient:exp-closed-form}.
    Hence, the optimal distribution $q$ must satisfy via said lemma that 
    \[
        \cond^{q}_{\config'}(\config) =  \frac{\exp(- \eta_{\nlvls} \genericloss(\config))}{\sum_{\config'' \in \children(\config')}  \exp(-\eta_{\nlvls} \genericloss(\config''))}
    \]
    and that
    \[
        \lan{\cond^{q}_{\config'},\genericloss} +  \frac{1}{\eta_{\nlvls}} \sum_{\config \in \children(\config')} \cond^{q}_{\config'}(\config) \ln \cond^{q}_{\config'}(\config) = - \frac{1}{\eta_{\nlvls}} \ln \lrb{ \sum_{\config \in \children(\config')} \exp(- \eta_{\nlvls} \genericloss(\config))} \,.
    \]
    Substituting back gives us that
    \begin{align*}
        &\lan{q,\genericloss} + \psi(q) = \sum_{\lvl \in [\nlvls-2]} \frac{1}{\eta_{\lvl}} \sum_{\config \in \expertree_\lvl} q[\config] \ln q[\config] + \bbrb{\frac{1}{\eta_{\nlvls-1}} + \frac{1}{\eta_{\nlvls}}} \sum_{\config \in \expertree_{\nlvls-1}} q[\config] \ln q[\config] \\
        &\hspace{15em}+ \sum_{\config' \in \expertree_{\nlvls-1}} q[\config'] \genericloss_{\nlvls-1}(\config') \,.
    \end{align*}
    Iterating this derivation upwards through the levels yields the lemma's statement.
\end{proof}

Combining \Cref{lem:efficient:exp-levels-closed-form} with the loss function decomposition described in \Cref{sec:efficient} yields the following result.

\begin{lemma} \label{lem:efficient:exp-profiles-closed-form}
    At round $t$, $\lvl \in \{0,\dots,\nlvls-1\}$, $\vertices \in \vertexset_{\lvl}$, and $a \in \actions_{\lvl}$, define
    \[
        \omg_{t, \vertices}(a) \coloneqq \sum_{\vertices' \in \children(\vertices)} -\etas_{\lvl+1} \ln \lrb{\sum_{b \in \suc_{\lvl+1}(a)} \exp\Brb{-\etas^{-1}_{\lvl+1} \brb{\closses_{t,\vertices'}(b,a) + \omg_{t,\vertices'}(b)}}} \,,
    \]
    while for $\vertices \in \vertexset_{\nlvls}$ and $a \in \actions_{\nlvls}$, set $\omg_{t,\vertices}(a) \coloneqq 0$. 
    Then, for $\config \in \expertree_\nlvls$,
    \begin{align*} %
        p_{t+1}(\config) =  \prod_{\vertices \in \vertexset_{1:\nlvls}} \cond_{t+1,\vertices}(\prof_{\vertices} \mid \prof_{\parent(\vertices)}) \,,
    \end{align*}
    where for $\vertices \in \vertexset_\lvl$ and $a \in \actions_{\lvl-1}$, $\cond_{t+1,\vertices}(\cdot \mid a)$ is a distribution over $\suc_{\lvl}(a)$ given by 
    \[
        \cond_{t+1,\vertices}(b \mid a) \propto \exp\Brb{- \etas^{-1}_{\lvl} \brb{\closses_{t,\vertices}(b,a) + \omg_{t,\vertices}(b)}} \,.
    \]
\end{lemma}
\begin{proof}
    Firstly, we extend the notation used in \Cref{lem:efficient:exp-levels-closed-form}.
    For $\config \in \expertree_{\nlvls}$, let $Y_{t,\nlvls}(\config) \coloneqq \sum_{s \leq t} \genloss_s(\config)$, while for $\lvl \in \{0,\dots,\nlvls - 1\}$ and $\config \in \expertree_{\lvl}$, define
    \[
        Y_{t,\lvl}(\config) \coloneqq - \etas_{\lvl+1} \ln \lrb{ \sum_{\config' \in \children(\config)} \exp\brb{- \etas^{-1}_{\lvl+1} Y_{t,\lvl+1}(\config')}} \,.
    \]
    For any $\config \in \expertree_{\nlvls-1}$,
    \begin{align*}
        &\sum_{\config' \in \children(\config)}\exp(-\etas^{-1}_{\nlvls} Y_{t,\nlvls}(\config'))  \\
        &\hspace{0em}= \sum_{\config' \in \children(\config)}\prod_{\vertices \in \vertexset_{1:\nlvls}} \exp\brb{-\etas^{-1}_{\nlvls} \closses_{t,\vertices}(\prof'_\vertices,\prof'_{\parent(\vertices)})}\\
        &\hspace{0em}= \prod_{\vertices \in \vertexset_{1:\nlvls-1}} \exp\brb{-\etas^{-1}_{\nlvls} \closses_{t,\vertices}(\prof_\vertices,\prof_{\parent(\vertices)})} \sum_{\config' \in \children(\config)}\prod_{\vertices \in \vertexset_{\nlvls}} \exp\brb{-\etas^{-1}_{\nlvls} \closses_{t,\vertices}(\prof'_\vertices,\prof_{\parent(\vertices)})}\\
        &\hspace{0em}= \prod_{\vertices \in \vertexset_{1:\nlvls-1}} \exp\brb{-\etas^{-1}_{\nlvls} \closses_{t,\vertices}(\prof_\vertices,\prof_{\parent(\vertices)})} \sum_{(b_\vertices)_{\vertices \in \vertexset_{\nlvls}} \in \bigtimes\limits_{\vertices \in \vertexset_{\nlvls}} \suc_{\nlvls}(\prof_{\parent(\vertices)})}\prod_{\vertices \in \vertexset_{\nlvls}} \exp\brb{-\etas^{-1}_{\nlvls} \closses_{t,\vertices}(b_\vertices,\prof_{\parent(\vertices)})}\\
        &\hspace{0em}= \prod_{\vertices \in \vertexset_{1:\nlvls-1}} \exp\brb{-\etas^{-1}_{\nlvls} \closses_{t,\vertices}(\prof_\vertices,\prof_{\parent(\vertices)})} \prod_{\vertices \in \vertexset_\nlvls} \sum_{b \in \suc_{\nlvls}(\prof_{\parent(\vertices)})} \exp\brb{-\etas^{-1}_{\nlvls} \closses_{t,\vertices}(b,\prof_{\parent(\vertices)})} \,,
    \end{align*}
    where the second equality uses that all children of $\config$ are identical up to and including level $\nlvls-1$; the third uses the definition of $\children(\config)$, and the last uses that the sum is separable over the elements of $\vertexset$.
    This now allows us to write
    \begin{align*}
        &Y_{t,\nlvls-1}(\config) \\
        &= - \etas_{\nlvls} \ln \lrb{ \sum_{\config' \in \children(\config)}\exp(-\etas^{-1}_{\nlvls} Y_{t,\nlvls}(\config')) }
        \\
        &= \sum_{\vertices \in \vertexset_{1:\nlvls-1}}  \closses_{t,\vertices}(\prof_\vertices,\prof_{\parent(\vertices)}) + \sum_{\vertices \in \vertexset_{\nlvls}} -\etas_{\nlvls} \ln \lrb{\sum_{b \in \suc_{\nlvls}(\prof_{\parent(\vertices)})} \exp\brb{-\etas^{-1}_{\nlvls} \closses_{t,\vertices}(b,\prof_{\parent(\vertices)})}}\\
        &= \sum_{\vertices \in \vertexset_{1:\nlvls-1}}  \closses_{t,\vertices}(\prof_\vertices,\prof_{\parent(\vertices)}) + \sum_{\vertices \in \vertexset_{\nlvls-1}} \sum_{\vertices' \in \children(\vertices)} -\etas_{\nlvls} \ln \lrb{\sum_{b \in \suc_{\nlvls}(\prof_{\vertices})} \exp\brb{-\etas^{-1}_{\nlvls} \closses_{t,\vertices'}(b,\prof_{\vertices})}} \\
        &= \sum_{\vertices \in \vertexset_{1:\nlvls-1}}  \closses_{t,\vertices}(\prof_\vertices,\prof_{\parent(\vertices)}) + \sum_{\vertices \in \vertexset_{\nlvls-1}} \omg_{t, \vertices}(\prof_\vertices) \\
        &= \sum_{\vertices \in \vertexset_{1:\nlvls-2}}  \closses_{t,\vertices}(\prof_\vertices,\prof_{\parent(\vertices)}) + \sum_{\vertices \in \vertexset_{\nlvls-1}} \brb{ \closses_{t,\vertices}(\prof_\vertices,\prof_{\parent(\vertices)}) + \omg_{t, \vertices}(\prof_\vertices)} \,,
    \end{align*}
    where the fourth equality follows from the definition of $\omg_{t,\vertices}$.
    With this, iterating the derivation above allows us to conclude that for $\lvl \in \{0,\dots,\nlvls\}$ and $\config\in \expertree_\lvl$, 
    \begin{align*}
        Y_{t,\lvl}(\config) = \sum_{\vertices \in \vertexset_{1:\lvl-1}}  \closses_{t,\vertices}(\prof_\vertices,\prof_{\parent(\vertices)}) + \sum_{\vertices \in \vertexset_{\lvl}} \brb{ \closses_{t,\vertices}(\prof_\vertices,\prof_{\parent(\vertices)}) + \omg_{t, \vertices}(\prof_\vertices)}  \,. 
    \end{align*}
    Then, thanks to  \Cref{lem:efficient:exp-levels-closed-form} and the definition of $p_t$, it holds for $\lvl \in \{0,\dots,\nlvls-1\}$ and $\config \in \expertree_\lvl$,
    \begin{align*}
    \cond^{p_{t+1}}_{\config}(\config') &= \frac{\exp\brb{- \etas^{-1}_{\lvl+1} Y_{t,\lvl+1}(\config')}}{  \exp\brb{-\etas^{-1}_{\lvl+1} Y_{t,\lvl}(\config)}} \\
    &= \frac{\prod_{\vertices \in \vertexset_{\lvl+1}}\exp\Brb{- \etas^{-1}_{\lvl+1} \brb{\closses_{t,\vertices}(\prof'_{\vertices},\prof_{\parent(\vertices)}) + \omg_{t,\vertices}(\prof'_{\vertices})}}}{\prod_{\vertices \in \vertexset_{\lvl}}\exp\brb{- \etas^{-1}_{\lvl+1} \omg_{t,\vertices}(\prof_{\vertices})}} \\
    &= \prod_{\vertices \in \vertexset_{\lvl+1}} \frac{\exp\Brb{- \etas^{-1}_{\lvl+1} \brb{\closses_{t,\vertices}(\prof'_{\vertices},\prof_{\parent(\vertices)}) + \omg_{t,\vertices}(\prof'_{\vertices})}}}{ \sum_{b \in \suc_{\lvl+1}(\prof_{\parent(\vertices)})} \exp\Brb{-\etas^{-1}_{\lvl+1} \brb{\closses_{t,\vertices}(b,\prof_{\parent(\vertices)}) + \omg_{t,\vertices}(b)}}} \quad \forall \config' \in \children(\config) \,,
    \end{align*}
    where in second step we cancelled the factors common to both the numerator and the denominator seeing as $\config' \in \children(\config)$, and the third step used the definition of $\omg_{t,\vertices}$.
    The sought result now follows after using that for any $\config \in \expertree_\nlvls$,
    \[
        p_{t+1}(\config) = \prod_{\lvl \in \{0,\dots,\nlvls-1\}} \cond_{\ancfun_{\lvl}(\config)}^{p_{t+1}}(\ancfun_{\lvl+1}(\config)) \,. 
    \]
\end{proof}

\subsection{Regret analysis}\label{app:efficient:regret}
The following is a proof of \Cref{thm:efficient}, which we restate below.
\thmefficient*
\begin{proof}
    As noted in \Cref{sec:efficient:alg}, \Cref{alg:efficient} mimics the sampling process of \Cref{alg:2-point-feedback-experts}.
    To be precise, the path $(\action_{t,\lvl})_{\lvl \leq \nlvls}$ plays the role of $(\anc{\varfunc_t}{\lvl}(x_t))_{\lvl \leq \nlvls}$, which gives indeed that $\bee_{t,1} = \action_{t,\lvl_t}$ and $\bee_{t,2} = \action_{t,\lvl_t-1}.$
    On the other hand, the path $(\action'_{t,\lvl})_{\lvl \leq \nlvls}$ plays the role of $(\anc{\func_{t,2}}{\lvl}(x_t))_{\lvl \leq \nlvls}$ as it branches off $(\action_{t,\lvl})_{\lvl \leq \nlvls}$ at level $\lvl_t -1$, hence why we set $A_{t,2} = \action'_{t,\nlvls}$.
    Finally, we do not sample a third path like \Cref{alg:2-point-feedback-experts}; instead, we use $(\action_{t,\lvl})_{\lvl \leq \nlvls}$ to also play the role of $(\anc{\func_{t,1}}{\lvl}(x_t))_{\lvl \leq \nlvls}$ and set $A_{t,1} = \action_{t,\nlvls}$.
    Taking $\func_{t,1} = \varfunc_t$ does not pose a problem in the analysis of \Cref{alg:2-point-feedback-experts} as long as $\varfunc_{t}(x_t)$ is independent of $\anc{\varfunc_t}{\lvl_t-1}(x_t)$ given $\lvl_t$ and $\anc{\varfunc_t}{\lvl_t}(x_t)$.
    This holds true here thanks to the factorized form of $p_t$.

    At every $\lvl$, it holds that $|\actions_\lvl| \leq 3^\adim 2^{\lvl\adim}$, $\max_{a \in \actions_{\lvl-1}} |\suc_{\lvl}(a)| = 4^\adim$, and $\ln |\expertree_\lvl| 
    \leq \ln(4) \adim 2^\cdim 2^{\lvl\cdim} $.
    Moreover, \Cref{lem:efficient:satisfying-assump} asserts that the tree satisfies \Cref{assump:function-tree} with $\apprxcst = \losslip$ and $\prtcst = 3 \cdot 2^{-2} \losslip$.
    Hence, \Cref{thm:generic-thm} with $\delta_\lvl \propto 2^{\lvl(\cdim+\adim-2)/3}$ and
    $
        \eta_{\lvl} = \frac{2^{\lvl}}{12 \losslip}\sqrt{ {\ln(4) \adim \delta_\lvl  2^{(\lvl+1)(\cdim-\adim)}}/( 3^\adim T ) }
    $
    gives that
     \begin{align*}
        \reg{T}{2} &\leq 192 \losslip \sqrt{\ln(4) \adim} \lrb{ \sum_{\lvl =1}^{\nlvls}    2^{(\lvl+3)(\adim+\cdim - 2)/3}   }^{3/2} \sqrt{T} + \losslip 2^{-\nlvls} T \\
        &\leq 192 \losslip \sqrt{\ln(4) \adim} \lrb{ \sum_{\lvl =1}^{\nlvls+3}    2^{\lvl(\adim+\cdim - 2)/3}   }^{3/2} \sqrt{T} + 8 \losslip 2^{-\nlvls-3} T \,.
    \end{align*}
    Similarly to the proof of \Cref{thm:concrete}, it holds that
    \begin{align*}
        \lrb{ \sum_{\lvl =1}^{\nlvls+3}    2^{\lvl(\adim+\cdim - 2)/3}   }^{3/2} 
        &\leq \begin{cases}
            11 \cdot 2^{(\adim + \cdim -2) (\nlvls+3)/2} &\text{if } \adim + \cdim > 2\\
            (\nlvls+3)^{3/2} &\text{if } \adim + \cdim = 2
        \end{cases} \,.
    \end{align*}
    So, setting $\nlvls =  \bfl{\frac{\log_2(T)}{\cdim + \adim } - 3}$ gives that 
    \begin{align*}
        \reg{T}{2} 
        \leq 16 \losslip \brb{1+ 132 \sqrt{\ln(4) \adim} } T^{1-\frac{1}{\adim + \cdim}}
    \end{align*}
    when $\adim + \cdim > 2$, and otherwise (when $\adim = \cdim=1$) yields that
    \begin{align*}
        \reg{T}{2} 
        \leq 16 \losslip \brb{1 + 12 \sqrt{\ln(4)} (\log_2(T)/2)^{3/2} } \sqrt{T}\,.
    \end{align*}
\end{proof}

\subsection{The predecessor set}\label{app:efficient:pre}
For every $\lvl \in\{0,\dots,\nlvls\}$, we define a function $\pre_{\lvl} \colon \actions_{\varlvl \geq \lvl} \rightarrow 2^{\actions_\lvl}$ given by $\pre_{\lvl}(b) \coloneqq \{b\}$ for $b \in \actions_\lvl$, by $\pre_{\lvl}(b) \coloneqq \{a \in \actions_\lvl \mid b \in \suc_{\lvl+1}(a)\}$ for $b \in \actions_{\lvl+1}$, and by
$
    \pre_{\lvl}(b) \coloneqq \bigcup_{a \in \pre_{\varlvl-1}(b)} \pre_{\lvl}(a) 
$
for $b \in \actions_\varlvl$ with $\varlvl \geq \lvl +2$.
From one level to the previous, this function is the reverse of the successor relation.
Going further up, we end up with the set of all actions that can lead to the action in question via a sequence of actions respecting the successor relation. 
The following result provides a uniform upper bound for $\pre_{\lvl}(\cdot)$. 
\begin{lemma}
    In the tree of \Cref{sec:efficient:tree}, it holds for any $\lvl \in [\nlvls]$ and $b \in \actions_{\varlvl \geq \lvl}$ that $|\pre_\lvl(b)| \leq  3^{\adim}$. 
\end{lemma}
\begin{proof}
    Fix some $\varlvl > \lvl$ and $b \in \actions_\varlvl$.
    Any $a \in \pre_{\varlvl-1}(b)$ satisfies that $\norm{a-b}_\infty \leq 3 \cdot 2^{-\varlvl-1}$.
    Therefore, $\pre_{\varlvl-1}(b)$ is contained in an $L_\infty$ ball of radius $3 \cdot 2^{-\varlvl-1}$ (or diameter $3 \cdot 2^{-\varlvl} = (3/2) \cdot 2^{-\varlvl+1}$) centered at $b$.
    At the same time, $\pre_{\varlvl-1}(b)$ is a subset of $\actions_{\varlvl-1}$ which is a $2^{-\varlvl+1}$ spaced grid.
    Hence, $|\pre_{\varlvl-1}(b)| \leq 2^\adim$.
    Moving upwards, $\pre_{\varlvl-2}(a)$ is contained in an $L_\infty$ ball of radius $3 \cdot 2^{-\varlvl}$ centered at $a$ for all $a\in\pre_{\varlvl-1}(b)$. 
    Hence, $\pre_{\varlvl-2}(b)$ is contained in an $L_\infty$ ball of diameter $9 \cdot 2^{-\varlvl} = ({9}/{4}) \cdot 2^{-\varlvl+2}$.
    This is now a subset of $\actions_{\varlvl-2}$, which is a $2^{-\varlvl+2}$ spaced grid, implying that $|\pre_{\varlvl-2}(b)| \leq 3^\adim$.
    Iterating this argument, we obtain a series $(y_{k})_{k=\varlvl-1,\dots,\lvl}$ with $y_{\varlvl-1} = 3/2$ and $y_{k} = (y_{k+1}+3)/2$, such that $\pre_{k}(b)$ is contained in an $L_\infty$ ball of diameter $y_k \cdot 2^{-k}$.
    Since for any $k$, $y_k < 3$ and $\actions_k$ is a $2^{-k}$ spaced grid, we obtain that $|\pre_{k}(b)| \leq 3^\adim$.
\end{proof}

\end{document}